\documentclass[journal]{IEEEtran}

\usepackage{amsmath,amsfonts,amsthm,amssymb,optidef,stmaryrd}
\usepackage{bm}
\usepackage{mdframed}
\usepackage{array}
\usepackage[caption=false,font=normalsize,labelfont=sf,textfont=sf]{subfig}
\usepackage{textcomp}
\usepackage{stfloats}
\usepackage{url}
\usepackage{verbatim}
\usepackage{graphicx}
\usepackage{cite}
\usepackage{color}
\usepackage{multirow,booktabs,makecell}
\usepackage[hidelinks]{hyperref}
\usepackage{enumitem}
\usepackage{siunitx}

\usepackage[linesnumbered,ruled,vlined]{algorithm2e}
\DontPrintSemicolon

\mdfsetup{%
innertopmargin=0pt,
innerleftmargin=3pt,
innerbottommargin=3pt,
innerrightmargin=3pt
}

\usepackage[all=normal,paragraphs=tight,floats=normal,mathspacing=normal,wordspacing=tight,charwidths=tight,mathdisplays=normal,leading=normal]{savetrees}

\definecolor{mygreen}{RGB}{11, 218, 81}
\definecolor{myblue}{RGB}{31, 81, 255}
\definecolor{mypurple}{RGB}{191, 64, 191}
\definecolor{myorange}{RGB}{255, 172, 28}
\definecolor{myred}{RGB}{220, 38, 38}

\newcommand{\abs}[1]{\ensuremath{\left| #1 \right|}}
\newcommand{\esp}[1]{\ensuremath{\mathbb{E}\left[ #1 \right]}}
\newcommand{\espo}[2]{\ensuremath{\mathbb{E}_{ #2 }\left[ #1 \right]}}
\newcommand{\norm}[2]{\ensuremath{\left\lVert #1 \right\rVert}_#2}

\newcommand{\tr}[1]{\ensuremath{\mathsf{Tr}\left( #1 \right)}}
\newcommand*{\transp}{{\mkern-1.5mu\mathsf{T}}}
\newcommand*{\herm}{{\mathsf{H}}}

\newcommand{\supp}[1]{\ensuremath{\textnormal{supp}\left( #1 \right)}}
\newcommand{\rk}[1]{\ensuremath{\textnormal{rank}\left( #1 \right)}}
\newcommand{\diag}[1]{\ensuremath{\textnormal{diag}\left( #1 \right)}}
\newcommand{\rowsupp}[1]{\ensuremath{\mathrm{rowsupp}\left( #1 \right)}}
\newcommand{\spanvec}[1]{\ensuremath{\textnormal{span}\left( #1 \right)}}

\newcommand*{\bs}{\mathbf{s}}

\newcommand*{\by}{\mathbf{y}}
\newcommand*{\bp}{\mathbf{p}}
\newcommand*{\bu}{\mathbf{u}}

\newcommand*{\bg}{\mathbf{g}}
\newcommand*{\ba}{\mathbf{a}}
\newcommand*{\be}{\mathbf{e}}
\newcommand*{\bn}{\mathbf{n}}

\newcommand*{\btheta}{\boldsymbol{\theta}}

\newcommand*{\bzeta}{\boldsymbol{\zeta}}

\newcommand*{\bgamma}{\boldsymbol{\gamma}}
\newcommand*{\bphi}{\boldsymbol{\phi}}
\newcommand*{\bomega}{\boldsymbol{\omega}}
\newcommand*{\bmu}{\boldsymbol{\mu}}
\newcommand*{\bsigma}{\boldsymbol{\sigma}}

\newcommand*{\bU}{\mathbf{U}}

\newcommand*{\bY}{\mathbf{Y}}
\newcommand*{\bA}{\mathbf{A}}
\newcommand*{\bB}{\mathbf{B}}
\newcommand*{\bR}{\mathbf{R}}
\newcommand*{\bI}{\mathbf{I}}
\newcommand*{\bS}{\mathbf{S}}
\newcommand*{\bN}{\mathbf{N}}
\newcommand*{\bZ}{\mathbf{Z}}

\newcommand*{\bP}{\mathbf{P}}

\newcommand*{\bSigma}{\mathbf{\Sigma}}
\newcommand*{\bLambda}{\mathbf{\Lambda}}
\newcommand*{\bPsi}{\mathbf{\Psi}}
\newcommand*{\bPi}{\mathbf{\Pi}}
\newcommand*{\bPhi}{\mathbf{\Phi}}

\newcommand*{\bGamma}{\mathbf{\Gamma}}

\newcommand*{\mcX}{\mathcal{X}}
\newcommand*{\mcY}{\mathcal{Y}}
\newcommand*{\mcZ}{\mathcal{Z}}

\newcommand*{\mcG}{\mathcal{G}}

\newcommand*{\mcM}{\mathcal{M}}

\newcommand*{\mcCN}{\mathcal{CN}}

\newcommand*{\mcV}{\mathcal{V}}

\newcommand*{\mcU}{\mathcal{U}}
\newcommand*{\mcB}{\mathcal{B}}

\newcommand*{\mcO}{\mathcal{O}}
\newcommand*{\mcP}{\mathcal{P}}
\newcommand*{\mcT}{\mathcal{T}}
\newcommand*{\mcE}{\mathcal{E}}
\newcommand*{\mcI}{\mathcal{I}}
\newcommand*{\mcA}{\mathcal{A}}
\newcommand*{\mcJ}{\mathcal{J}}
\newcommand*{\mcW}{\mathcal{W}}

\newcommand*{\mbbC}{\mathbb{C}}
\newcommand*{\mbbR}{\mathbb{R}}

\newcommand*{\mbbS}{\mathbb{S}}

\usepackage{xstring} 

\newcommand{\ej}[2][+]{%
    \ensuremath{\mathrm{e}^{%
        \IfStrEq{#1}{-}{-\mathrm{j}}{\mathrm{j}} #2 }%
    }}%

\DeclareMathOperator*{\argmax}{arg\,max}

\newtheorem{remark}{Remark}[section]
\newtheorem{proposition}{Proposition}
\newtheorem{definition}{Definition}

\newcommand*{\stilde}[1]{\vphantom{#1}\smash{\tilde{#1}}}
\newcommand*{\shat}[1]{\vphantom{#1}\smash{\hat{#1}}}

\begin{document}

\title{Model-based learning for DoA estimation and array calibration via continuous sparse recovery}

\author{
	Baptiste Chatelier, 
    Jos\'{e} Miguel Mateos-Ramos,
    Vincent Corlay,
    Matthieu Crussière,\\
    Christian H\"{a}ger,
    Henk Wymeersch,
    Nir Shlezinger,
    Luc Le Magoarou
    \thanks{This work was supported by the French national research agency (MoBAIWL project, grant ANR-23-CE25-0013), and by the SNS JU project 6G-DISAC funded under the EU's Horizon Europe research and innovation program under Grant Agreement No. 101139130. Computational resources were provided by the National Academic Infrastructure for Supercomputing in Sweden (NAISS), funded by the Swedish Research Council.}
    \thanks{Baptiste Chatelier is with INSA Rennes, CNRS, IETR-UMR 6164, Rennes, France and the Department of Electrical Engineering, Chalmers University of Technology, Sweden (email: bapcha@chalmers.se).}
    \thanks{Matthieu Crussière and Luc Le Magoarou are with INSA Rennes, CNRS, IETR-UMR 6164, Rennes, France (email: \{matthieu.crussiere ; luc.le-magoarou\}@insa-rennes.fr).}
    \thanks{Jos\'{e} Miguel Mateos-Ramos, Christian H\"{a}ger and Henk Wymeersch are with the Department of Electrical Engineering, Chalmers University of Technology, Sweden (email: \{josemi ; christian.haeger ; henkw\}@chalmers.se).}%
    \thanks{Vincent Corlay is with Mitsubishi Electric R\&D Centre Europe, Rennes, France (email: v.corlay@fr.merce.mee.com).}
    \thanks{Nir Shlezinger is with the School of ECE, Ben-Gurion University of the Negev, Be'er-Sheva, Israel (email: nirshl@bgu.ac.il).}
	}



\maketitle

\begin{abstract}

    Direction of arrival (DoA) estimation, a classical signal processing problem, has experienced renewed interest with the advent of integrated sensing and communication (ISAC). While decades of research have yielded highly accurate DoA estimators, hardware impairments have frequently been neglected, despite degrading both DoA estimation accuracy and downstream communication performance in ISAC systems. This work introduces a joint array calibration and DoA estimation method based on the simultaneous orthogonal matching pursuit (SOMP) algorithm, following the model-based machine learning paradigm. The proposed method enables the calibration of antenna arrays subject to severe gain-phase and antenna location impairments through projected gradient descent (PGD), with either supervised or unsupervised objectives. Additionally, DoA estimation is enhanced by extending SOMP to continuous dictionaries via PGD refinement steps. Simulation results demonstrate that the proposed method enables efficient hardware impairment compensation, even without labeled samples, while outperforming conventional sparse recovery and subspace-based DoA estimation baselines.

\end{abstract}

\begin{IEEEkeywords}
    DoA estimation, Hardware impairments, Model-based machine learning, Sparse recovery, ISAC.
\end{IEEEkeywords}

\section{Introduction}\label{sec:introduction}

\IEEEPARstart{D}{irection} of arrival (DoA) estimation refers to the process of determining the angle at which a propagating wave impinges upon an antenna array. While pioneering work on DoA estimation can be traced back to the 1940s, the topic has experienced renewed interest within the wireless communication research community with the advent of integrated sensing and communication (ISAC) systems. In such systems, sensed information, such as DoAs, is leveraged to increase performance of wireless transmission methods, rendering efficient and accurate sensing a problem of high practical interest.

Besides beamforming (e.g. Bartlett or Capon~\cite[Section~6.3]{Stoica2005}), classical model-based DoA estimation methods broadly fall into two categories: subspace methods~\cite{Schmidt1986,Roy1989,Stoica2005} and sparse recovery approaches~\cite{Yang2018}. The subspace approach exploits the eigendecomposition of the signal received by the antenna array, with well-known representatives including the multiple signal classification (MUSIC)~\cite{Schmidt1986} and estimation of signal parameters via rotational invariance (ESPRIT)~\cite{Roy1989} algorithms. The sparse recovery approach, on the other hand, exploits the sparsity of the received signal in the angular domain, e.g. using greedy pursuit algorithms such as orthogonal matching pursuit (OMP)~\cite{Mallat1993,Pati1993}. Despite their strong performance, these methods have two main limitations: subspace methods degrade significantly in the presence of coherent sources, and both classes of methods remain sensitive to model mismatches induced by hardware impairments (HWIs). Recently, several studies have investigated the impact of HWIs in both sensing~\cite{Bozorgi21,Tubail24,Chen24} and communication systems~\cite{Bjornson13,Bjornson14,Bjornson15,Hohne10,chatelier_impact_2022}, motivating the development of compensation strategies. However, model-based calibration strategies typically rely on known impairment models or require reference signals with known parameters, e.g. ground-truth source positions, which are rarely available in practice.

These limitations have motivated growing interest in data-driven alternatives, spurred by recent advances in machine learning (ML). By learning directly from observed data, such methods can compensate for model mismatches, and have already been applied to DoA estimation, with multilayer perceptrons in~\cite{Chen20,Cong21} or convolutional neural networks (CNNs) in~\cite{Papageorgiou21,Zheng24}. Nevertheless, they are subject to inherent limitations: first, they lack the interpretability of classical model-based approaches and are often regarded as black boxes; second, they typically require extensive data collection and considerable training overhead.

To address these limitations, the model-based machine learning (MB-ML) paradigm~\cite{Shlezinger23} combines both approaches, incorporating domain knowledge from signal processing to structure, initialize or optimize ML methods. The objective is twofold: enhance the interpretability and reduce the complexity of ML methods, while mitigating the model bias inherent to signal processing approaches. This paradigm has recently shown promising results in several wireless communication problems, including channel estimation~\cite{Yassine22,Chatelier23,Klaimi26,Chatelier2025_loc2chan}, filter design~\cite{Revach2021,Ni24,Shlezinger25}, ISAC system design~\cite{mateos2023model,mateosramos2025,Nguyen25,mateosramos2026} and localization~\cite{Chatelier2025_loc,Gast2026}. It has also been applied specifically to DoA estimation with deep-learning aided MUSIC algorithms~\cite{Elbir20}, covariance reconstruction networks~\cite{Shiran26} and surrogate covariance learning approaches such as SubspaceNet~\cite{Shmuel2025} and its variants~\cite{Gast2025,Gast2026}.

Given the inherent limitations of classical model-based methods, and the reliance of learning-based approaches on labeled calibration data when HWIs are considered, two questions naturally arise: \textit{Can the MB-ML paradigm be leveraged to address the joint array calibration and DoA estimation problem in the presence of HWIs without labeled data? How does the computational complexity of the resulting method compare to that of classical approaches?}

\noindent \textbf{Contributions.} This work focuses on the application of the MB-ML paradigm to the joint DoA estimation and array calibration problem within the sparse recovery framework. More specifically, this joint estimation problem is defined over two sets of optimization variables: the DoAs to be estimated and the array parameters to be calibrated. This problem is then addressed sequentially, first calibrating the HWIs and subsequently estimating the DoAs. The proposed method leverages the simultaneous orthogonal matching pursuit (SOMP) algorithm~\cite{Tropp06} as a coarse initialization for both the calibration and estimation stages, which are subsequently refined via first-order optimization methods. The main contributions of this paper are as follows.

\begin{itemize}
    \item \textbf{Unsupervised calibration via dictionary refinement.} Since the atoms of the SOMP dictionary depend on the array parameters, which are perturbed by HWIs, calibration is cast as a dictionary refinement problem. SOMP first selects the active atoms from multiple snapshots. Then, projected gradient descent (PGD) iteratively updates the array parameters over the resulting reduced dictionary through an unsupervised objective.
    \item \textbf{Off-grid estimation.} Once the array is calibrated, SOMP over a coarse dictionary provides initial DoA estimates, which are then refined by PGD in the continuous angular domain under an unsupervised objective. This refinement decouples estimation accuracy from the dictionary discretization.
    \item \textbf{Complexity analysis.} The computational complexity of the proposed calibration and estimation methods is derived and compared with that of MUSIC and SOMP.
    \item \textbf{Validation.} Extensive experiments show that the proposed method compensates for severe gain-phase and antenna location HWIs, and accurately estimates DoAs in low-snapshot, low-signal-to-noise ratio (SNR) and coherent-source regimes.
\end{itemize}

\noindent \textbf{Related work.} As previously mentioned, the MB-ML paradigm has been recently applied to the DoA estimation problem in~\cite{Shmuel2025,Gast2025,Chatelier2025,Gast2026}. In~\cite{Shmuel2025,Gast2025,Gast2026}, a CNN is trained to learn a surrogate covariance matrix, which is then fed to MUSIC or ESPRIT to mitigate the impact of HWIs and coherent sources on subspace methods. While this approach performs well in both the far-field and near-field regimes, and for coherent sources, it relies on supervised training with labeled DoAs, and the large number of CNN parameters incurs a significant computational cost.

A differentiable MUSIC algorithm was proposed in our previous work~\cite{Chatelier2025} for joint HWI calibration and DoA estimation. This method directly learns the physical parameters affected by HWIs, i.e. antenna locations and gains, and thus involves only a few learnable parameters. The recent work~\cite{Merkofer2026} adopts a similar strategy but directly learns the array manifold distortion. This improves generality with respect to the distortion sources, at the cost of interpretability, and relies on supervised training, whereas the method in~\cite{Chatelier2025} can operate with an unsupervised objective. Nevertheless, both approaches are inapplicable to coherent sources, as subspace methods rely on a full-rank source covariance matrix. This paper therefore departs from the subspace paradigm and considers sparse recovery techniques, which are naturally better suited to this setting.

The sparse recovery approach to joint HWI compensation and DoA estimation has been explored in~\cite{Liu2016}. While the authors account for gain-phase, antenna location, and mutual coupling impairments, the impact of these impairments on the atom structure is not fully exploited. Specifically, in the presence of antenna location impairments, rather than learning a per-antenna location correction term, as proposed in this paper, one correction term per antenna and DoA has to be estimated, resulting in a substantially larger number of parameters to optimize.

A closely related approach was proposed in the preprint~\cite{Ghanbari2017}, where SOMP is combined with gradient-descent updates to jointly estimate the DoAs and compensate for antenna gain-phase and mutual coupling impairments. However, antenna location impairments are not considered. Moreover, the entire SOMP dictionary is treated as an optimization variable, and SOMP is re-run at every gradient iteration. In contrast, the estimation method proposed here calls SOMP only once for initialization, and subsequently refine a reduced dictionary through a small set of physical parameters. Finally, the validation in~\cite{Ghanbari2017} is restricted to a single DoA configuration with non-coherent sources, whereas this work considers multiple DoA and HWI realizations as well as coherent sources.


\noindent \textbf{Organization.} The rest of the paper is organized as follows. Section~\ref{sec:system_model} presents the system model and the generic joint DoA estimation and array calibration problem. Section~\ref{sec:classical} recalls both classical subspace and sparse recovery methods. Section~\ref{sec:proposed} introduces the proposed MB-ML method and presents its computational complexity, and Section~\ref{sec:experiments} evaluates its performance against several baselines in different scenarios. Finally, Section~\ref{sec:conclusion} presents some concluding remarks as well as future research directions.

\noindent \textbf{Notations.} Lowercase bold letters represent vectors while uppercase bold letters represent matrices. $\mbbR$ and $\mbbC$ respectively denote the sets of real and complex numbers, and, given two integers $a \leq b$, $\llbracket a, b \rrbracket$ denotes the set of integers from $a$ to $b$. $\mcU\left(\mbbS\right)$ denotes the uniform distribution of the set $\mbbS$, $\mcCN\left(\bmu,\bSigma\right)$ denotes the complex Gaussian distribution of mean $\bmu$ and covariance matrix $\bSigma$. $\left(\cdot\right)^\transp$, $\left(\cdot\right)^\herm$, $\left(\cdot\right)^\dagger$ and $\rk{\cdot}$ respectively denote the transpose, Hermitian, Moore-Penrose pseudo-inverse and rank matrix operators. $\norm{\cdot}{p}$ and $\norm{\cdot}{\mathsf{F}}$ respectively denote the $\ell_p$ and Frobenius norms. $\abs{\cdot}$ denotes the absolute value for real numbers, modulus for complex numbers and cardinality for sets. $\mcO\left(\cdot\right)$ denotes the big-O Bachmann-Landau notation.

\section{System model and problem formulation}\label{sec:system_model}

\subsection{System model}

The considered sensing system comprises a uniform linear array (ULA) of $N$ elements with half-wavelength inter-element spacing, on which $M<N$ far-field sources impinge.\footnote{Note that although the system model is presented for a ULA, it extends straightforwardly to arbitrary array geometries.} Let $\mcV_{\Omega} = \left[ -\pi/2,  \pi/2 \right]$ denote the source angular domain. At time $t$, the signal measured by the antenna array, $\by\left(t\right) \in \mbbC^N$, can be expressed as:
\begin{equation}\label{eq:received_signal}
    \by\left(t\right) = \sum_{m=1}^M \ba_{\bzeta}\left(\theta_m \right) s_m\left(t\right) + \bn\left(t\right),
\end{equation}
where $\btheta = \left[\theta_1, \cdots, \theta_M\right]^\transp \in \mcV_{\Omega}^M$ denotes the source DoAs and $\ba_{\bzeta}\left(\theta\right) \in \mbbC^N$ is the steering vector representing the array response to a source with DoA $\theta$:
\begin{equation}\label{eq:steering_vector}
    \ba_{\bzeta}\left(\theta\right) = \frac{1}{\norm{\bg}{2}} \begin{bmatrix} g_1 \ej[-]{\frac{2\pi}{\lambda} p_1 \sin\left(\theta\right) } \\
    \vdots\\
    g_N \ej[-]{\frac{2\pi}{\lambda} p_N \sin\left(\theta\right)} \end{bmatrix},
\end{equation}
where $\bp = \left[p_1, \cdots, p_N\right]^\transp \in \mbbR^N$ and $\bg = \left[g_1, \cdots, g_N\right]^\transp \in \mbbC^N$ respectively denote the antenna locations and gains. $\bzeta = \left[\bp,\bg\right] \in \mcZ \subset \mbbR^N \times \mbbC^N$ denotes the array manifold parametrization, i.e. the physical parameters describing the array manifold $\mcM_{\bzeta} = \left\{\ba_{\bzeta}\left(\theta\right) \left\vert \theta \in \mcV_{\Omega} \right.\right\} \subset \mbbC^N$. Finally, $\bn\left(t\right) \in \mbbC^N \sim \mcCN\left(\mathbf{0}_N,\sigma^2_n \mathbf{I}_N\right)$ denotes the sensing noise, and $s_m\left(t\right) \in \mbbC$ represents the $m$th source signal.

It is considered that the sensing system captures $T$ snapshots, so that Eq.~\eqref{eq:received_signal} rewrites as:
\begin{equation}\label{eq:received_signal_matrix}
    \bY = \bA_{\bzeta}\left(\btheta\right) \bS + \bN,
\end{equation}
where $\bY = \left[\by\left(1\right), \cdots, \by\left(T\right)\right] \in \mbbC^{N\times T}$ is the sensed signal, $\bA_{\bzeta}\left(\btheta\right) = \left[ \ba_{\bzeta}\left(\theta_1\right), \cdots, \ba_{\bzeta}\left(\theta_M\right)\right] \in \mbbC^{N\times M}$ denotes the steering matrix, $\bS = \left[\bs\left(1\right), \cdots, \bs\left(T\right)\right] \in \mbbC^{M \times T}$ denotes the source signals with $\bs\left(i\right) = \left[s_1\left(i\right), \cdots, s_M\left(i\right) \right]^\transp \in \mbbC^M$, and $\bN = \left[\bn(1), \cdots \bn\left(T\right)\right] \in \mbbC^{N \times T}$ denotes the sensing noise.

\subsection{Problem formulation}
In practice, the true array manifold parametrization $\bzeta$ is not known, due to HWIs. Indeed, manufacturing errors may result in an antenna array whose inter-element spacings deviate from the nominal half-wavelength, and whose complex gains differ from one antenna to another. This imperfect knowledge has a direct impact on the system model formulation, as the steering vectors of Eq.~\eqref{eq:received_signal} and~\eqref{eq:received_signal_matrix} are then evaluated with a mismatched parametrization, thereby inducing errors when leveraging these models to perform DoA estimation.

In this setting, the generic joint DoA/HWI estimation problem can be formally defined as the following optimization problem:
\begin{mini!}
    {\psi, \bzeta}{\espo{\mu\left(\btheta, \psi\left(\bY ; \bzeta\right)\right)}{\left(\btheta, \bY \right) \sim p_{\btheta,\bY}}, \tag{$\mathsf{P1}$}\label{prob:optim_problem}}
    {}{}
    \addConstraint{\psi \in \mcE}{\nonumber}
    \addConstraint{\bzeta \in \mcZ}{\nonumber}
\end{mini!}
where:
\begin{itemize}
    \item $\psi \colon \mbbC^{N\times T} \rightarrow \mcV_{\Omega}^M$ is a DoA estimator from the generic estimator class $\mcE$.
    \item $\bzeta$ is the array manifold parametrization from the feasible set $\mcZ \subset \mbbR^N \times \mbbC^N$.
    \item $\btheta, \bY$ represent the true DoAs and sensed signals, following joint distribution $p_{\btheta,\bY}$.
    \item $\mu \colon \mcV_{\Omega}^M \times \mcV_{\Omega}^M \rightarrow \mbbR^+$ denotes an angular error measure.
\end{itemize}
Since this problem is intractable in its current form, this work aims to provide an approximate solution to~\eqref{prob:optim_problem} with satisfactory estimation performance and limited computational complexity.

\section{Classical methods}\label{sec:classical}

This section reviews the classical subspace and sparse recovery methods for DoA estimation. As discussed in~\cite{Shmuel2025}, these methods rely on some of the following assumptions, whose implications are presented in Section~\ref{subsec:limits}.

\begin{enumerate}[leftmargin=*,label=\textcolor{myblue}{(\textsf{A\arabic*})}]
    \item The sources are narrowband such that Eq.~\eqref{eq:received_signal} holds.\label{constraint:narrowband}
    \item The array manifold parametrization $\bzeta$ is perfectly known.\label{constraint:full_knowledge}
    \item The sources are non-coherent, i.e. $\bS$ is full rank.\label{constraint:non_coh}
    \item The number of snapshots $T$ is sufficiently high.\label{constraint:snapshots}
    \item The SNR is sufficiently high.\label{constraint:snr}
    \item The DoAs are estimated on a discrete grid.\label{constraint:discrete}
\end{enumerate}

\subsection{Subspace methods}

Subspace methods~\cite[Chapter~6]{Stoica2005} rely on the eigenstructure of the received signal covariance matrix. Indeed, the latter is expressed as:
\begin{equation}
    \bR_{\bY} = \esp{\by\left(t\right)\by\left(t\right)^\herm} =\bA_{\bzeta}\left(\btheta\right) \bR_{\bS} \bA^\herm_{\bzeta}\left(\btheta\right) + \sigma^2_n \mathbf{I}_N,
\end{equation}
where the covariance matrix of the source signals $\bR_{\bS} = \mathbb{E}[\bs\left(t\right)\bs\left(t\right)^\herm] = \sigma^2_s \mathbf{I}_M$ when~\ref{constraint:non_coh} and equal source powers are assumed. Then, considering that the source DoAs are distinct, one obtains:
\begin{equation}
    \rk{\bA_{\bzeta}\left(\btheta\right) \bR_{\bS} \bA^\herm_{\bzeta}\left(\btheta\right)} = \rk{\bR_{\bS}} = M,
\end{equation}
so that the eigenvalue decomposition (EVD) of $\bA_{\bzeta}\left(\btheta\right) \bR_{\bS} \bA^\herm_{\bzeta}\left(\btheta\right)$ admits only $M$ non-null eigenvalues. Remarking that $\bA_{\bzeta}\left(\btheta\right) \bR_{\bS} \bA^\herm_{\bzeta}\left(\btheta\right) = \sigma^2_s \bA_{\bzeta}\left(\btheta\right)\bA^\herm_{\bzeta}\left(\btheta\right)$ is Hermitian, it admits the following EVD:
\begin{equation}
    \sigma^2_s \bA_{\bzeta}\left(\btheta\right)\bA^\herm_{\bzeta}\left(\btheta\right) = \sigma^2_s \bU \bLambda_{\mathsf{S}}\bU^\herm,
\end{equation}
where $\bLambda_{\mathsf{S}} = \diag{\stilde{\lambda}_1, \cdots, \stilde{\lambda}_M, 0, \cdots, 0} \in \mbbR^{N \times N}$ denotes a diagonal matrix of eigenvalues, and $\bU \in \mbbC^{N \times N}$ is the corresponding eigenvector matrix, with $\bU\bU^\herm = \mathbf{I}_N$. Then one can rewrite $\bR_{\bY} = \bU \bLambda\bU^\herm$ with $\bLambda = \diag{\sigma^2_s \stilde{\lambda}_1 + \sigma^2_n, \cdots, \sigma^2_s \stilde{\lambda}_M + \sigma^2_n, \sigma^2_n, \cdots, \sigma^2_n} \in \mbbR^{N \times N}$. As this defines the EVD of $\bR_{\bY}$, the following comments regarding its eigenstructure can be made:
\begin{enumerate}
    \item The eigenvalues of $\bR_{\bY}$, ordered by decreasing amplitude, are $\sigma^2_s \stilde{\lambda}_1 + \sigma^2_n, \cdots, \sigma^2_s \stilde{\lambda}_M + \sigma^2_n, \sigma^2_n, \cdots, \sigma^2_n$, where $\sigma^2_n$ has multiplicity $N-M$.
    \item The eigenvectors can be partitioned into signal and noise subspace bases i.e. $\bU = \left[\bU_{\mathsf{S}}, \bU_{\mathsf{N}}\right]$, such that $\spanvec{\bU_{\mathsf{S}}} \perp \spanvec{\bU_{\mathsf{N}}}$.
\end{enumerate}

This particular eigenstructure is leveraged by subspace methods to perform DoA estimation. In particular, the MUSIC algorithm~\cite{Schmidt1986} builds on the orthogonality between signal and noise subspaces. Indeed, since $\spanvec{\bU_{\mathsf{S}}} = \spanvec{\bA_{\bzeta}\left(\btheta\right)} \Rightarrow \spanvec{\bA_{\bzeta}\left(\btheta\right)} \perp \spanvec{\bU_{\mathsf{N}}}$, it follows that:
\begin{equation}\label{eq:music_orthogonality}
    \forall i \in \llbracket 1, M \rrbracket, \norm{\bU^{\herm}_{\mathsf{N}} \ba_{\bzeta}\left(\theta_i\right)}{2}^2 = 0.
\end{equation}
MUSIC exploits this property to construct a spectrum whose peaks are located near the source DoAs, evaluated over a discretized angular domain $\mcV_{\Omega_d} \subset \mcV_{\Omega}$, as described in Algorithm~\ref{algo:MUSIC}. The DoAs are then estimated as the arguments of the $M$ highest-valued peaks, returned by the $\mathrm{peaks}_M$ operator.

\begin{algorithm}[t]
    \caption{$\mathtt{MUSIC}\left(\bY;\mcV_{\Omega_d},\bzeta\right)$}\label{algo:MUSIC}

    \KwIn{Received signal $\bY$, discrete angular domain $\mcV_{\Omega_d}$,\\ array manifold parametrization $\bzeta$}

    \tcp{Compute sample covariance matrix}
    $\hat{\bR}_{\bY} \gets \frac{1}{T} \bY \bY^\herm$\;

    \tcp{Perform EVD and subspace separation}
    $\left(\hat{\bU}_{\mathsf{S}},\hat{\bU}_{\mathsf{N}},\hat{\bLambda}\right) \gets \mathrm{EVD}\left(\hat{\bR}_{\bY}\right)$\;

    \tcp{Compute MUSIC spectrum}
    \ForEach{$\stilde{\theta} \in \mcV_{\Omega_d}$}{
        $P_{\mathrm{MUSIC}}\left(\stilde{\theta}; \bzeta, \bY\right) \gets \norm{\hat{\bU}^\herm_{\mathsf{N}} \ba_{\bzeta}\left(\stilde{\theta}\right)}{2}^{-2}$\;
    }

    \tcp{Estimate DoAs}
    $\hat{\btheta} \gets \underset{\stilde{\theta} \in \mcV_{\Omega_d}}{\mathrm{peaks}_M}\ P_{\mathrm{MUSIC}}\left(\stilde{\theta}; \bzeta, \bY\right)$\;

    \KwOut{Estimated DoAs $\hat{\btheta}$}
\end{algorithm}

\subsection{Sparse recovery methods}

\noindent\textbf{Continuous angular domain.} An alternative approach to DoA estimation consists in leveraging the linear combination structure of the received signal in Eq.~\eqref{eq:received_signal_matrix}. Since $\bY$ is a combination of only $M$ steering vectors drawn from the array manifold $\mcM_{\bzeta}$, DoA estimation can be cast as a sparse recovery problem over a continuously parametrized dictionary. From a generic perspective, this amounts to solving a sparse recovery problem with an atomic norm constraint~\cite[Section~11.6, pp.~557-571]{Yang2018}\cite{Tang13}, as detailed below.

\begin{definition}[{$\ell_0$ atomic norm}]\label{def:l0_atomic_norm}
    Given the steering vectors defined in Eq.~\eqref{eq:steering_vector}, the atomic set induced by the array manifold parametrization $\bzeta$ is defined as~\cite[Section~11.6.5, p.~558]{Yang2018}:
    \begin{equation}
        \mcA = \left\{\ba_{\bzeta}\left(\theta\right) \bphi^\transp \left\vert \theta \in \mcV_{\Omega}, \bphi \in \mbbC^T \textnormal{ s.t. }\norm{\bphi}{2}=1 \right.\right\}.
    \end{equation}
    Then, denoting $\bA_{\bzeta,k} = \ba_{\bzeta}\left(\theta_k\right)\bphi_k^\transp \in \mcA$, the $\ell_0$ atomic norm induced by $\mcA$ is expressed as:
    \begin{align}\label{eq:l0_atomic_norm}
        \norm{\bY}{{\mcA,0}} = \inf \left\{K \left\vert \bY = \sum_{k=1}^K c_k \bA_{\bzeta,k}, c_k > 0, \bA_{\bzeta,k} \in \mcA \right.\right\}.
    \end{align}
\end{definition}

\begin{remark}[Interpretation of the $\ell_0$ atomic norm]
    Intuitively, $\norm{\bY}{{\mcA,0}}$ represents the minimum number of rank-one contributions built from atoms required to represent $\bY$ in $\mcA$. Indeed, in the noiseless case, denoting by $\tilde{\bs}_m \in \mbbC^{T}$ the $m$th row of $\bS$, Eq.~\eqref{eq:received_signal_matrix} rewrites as:
    \begin{equation}
        \bY = \sum_{m=1}^M \norm{\tilde{\bs}_m}{2} \ba_{\bzeta}\left(\theta_m\right)\frac{\tilde{\bs}_m^\transp}{\norm{\tilde{\bs}_m}{2}},
    \end{equation}
    so that $\norm{\bY}{{\mcA,0}} \leq M$, with equality whenever the DoAs $\theta_m$ are distinct. The $\ell_0$ atomic norm hence recovers the number of sources.
\end{remark}

\begin{definition}[{Atomic support}]\label{def:atomic_support}
    Denoting a decomposition $\bY = \sum_{k=1}^{K^\star} c_k \ba_{\bzeta}\left(\theta_k\right)\bphi_k^\transp$ achieving $K^\star = \norm{\bY}{{\mcA,0}}$, the atomic support of $\bY$ is defined as the DoAs appearing in this decomposition:
    \begin{equation}\label{eq:atomic_support}
        \supp{\bY} = \left[\theta_1, \cdots, \theta_{K^\star}\right] \in \mcV_{\Omega}^{K^\star}.
    \end{equation}
\end{definition}

Following Definition~\ref{def:l0_atomic_norm}, for a given received signal $\bY$ and a known number of sources $M$, DoA estimation can be expressed as a sparse recovery problem using the atomic set $\mcA$:
\begin{mini!}
    {\bZ \in \mbbC^{N \times T}}{\norm{\bY - \bZ}{\mathsf{F}}^2.\tag{$\mathsf{P2}$}\label{prob:sparse_rec_atomic_norm}}
    {}{}
    \addConstraint{\norm{\bZ}{{\mcA,0}} \leq M}{\nonumber}
\end{mini!}
Given a solution $\bZ^\star$ of~\eqref{prob:sparse_rec_atomic_norm} and leveraging Definition~\ref{def:atomic_support}, the DoA estimates are given by $\hat{\btheta} = \supp{\bZ^\star}$. Solving~\eqref{prob:sparse_rec_atomic_norm} is however not straightforward as $\norm{\cdot}{{\mcA,0}}$ is non-convex and its evaluation is NP-hard in general. A convex surrogate is given by the atomic norm $\norm{\cdot}{\mcA}$, defined as the gauge function of the convex hull of $\mcA$. However, this relaxation is also generally not computationally tractable. 

\begin{remark}[SDP formulation]
    For a ULA without any antenna location impairments, and under~\ref{constraint:full_knowledge}, the steering vectors exhibit a Vandermonde structure, and $\norm{\cdot}{\mcA}$ therefore admits an equivalent semidefinite programming (SDP) formulation~\cite[Theorem~11.14]{Yang2018}~\cite{Tang13}. However, this structure is lost in the presence of antenna location impairments, and solving this SDP is in any case computationally intensive~\cite[Section~11.6.8, p.~568]{Yang2018}.
\end{remark}

Overall, the complexity limitations of atomic norm minimization motivate discretizing the angular domain $\mcV_{\Omega}$ into a finite grid, over which efficient \textit{greedy sparse recovery algorithms} can be applied, as detailed below. This discretization is however precisely what introduces~\ref{constraint:discrete}, whose impact is discussed in Section~\ref{subsec:limits}.

\noindent\textbf{Discretized angular domain.} In the multiple snapshot case, given a discretized angular domain $\mcV_{\Omega_d} = \left\{\stilde{\theta}_1, \cdots, \stilde{\theta}_A\right\} \subset \mcV_{\Omega}$ and a dictionary $\bPhi_{\bzeta} = \left[\bphi_{\bzeta,1}, \cdots, \bphi_{\bzeta,A}\right] \in \mbbC^{N \times A}$, whose columns are steering vectors evaluated on $\mcV_{\Omega_d}$, i.e. $\bphi_{\bzeta,k} = \ba_{\bzeta}\left(\stilde{\theta}_k\right)$, the received signal can be approximately decomposed as:
\begin{equation}\label{eq:discretized_decomposition}
    \bY \simeq \bPhi_{\bzeta} \bGamma = \sum_{k=1}^{A} \bphi_{\bzeta,k}\tilde{\bgamma}_k^\transp,
\end{equation}
where $\bGamma \in \mbbC^{A \times T}$ and $\tilde{\bgamma}_k \in \mbbC^{T}$ respectively denote a coefficient matrix and its $k$th row.

\begin{definition}[{Discrete atomic support}]
    Given the coefficient matrix $\bGamma \in \mbbC^{A \times T}$ in Eq.~\eqref{eq:discretized_decomposition}, the row support is defined as:
    \begin{equation}
        \rowsupp{\bGamma} = \left\{k \in \llbracket 1,A \rrbracket \left\vert \exists t \in \llbracket 1,T \rrbracket \textnormal{ s.t. } \gamma_{k,t} \neq 0 \right.\right\}.
    \end{equation}
    It defines the $\textnormal{row}-\ell_0$ quasi-norm:
    \begin{equation}
        \norm{\bGamma}{{\mathrm{row}-0}} = \abs{\rowsupp{\bGamma}},
    \end{equation}
    as well as the discretized equivalent of the atomic support:
    \begin{equation}
        \supp{\bPhi_{\bzeta}\bGamma} = \left[\stilde{\theta}_k \right]_{k \in \rowsupp{\bGamma}},
    \end{equation}
    where $\forall k,\ \stilde{\theta}_k \in \mcV_{\Omega_d}$.
\end{definition}

In this discretized setting, sparse recovery is performed over the discrete dictionary $\bPhi_{\bzeta}$:
\begin{mini!}
    {\bGamma \in \mbbC^{A \times T}}{\norm{\bY - \bPhi_{\bzeta}\bGamma}{\mathsf{F}}^2,\tag{$\mathsf{P3}$}\label{prob:sparse_rec_discretized}}
    {}{}
    \addConstraint{\norm{\bGamma}{{\mathrm{row}-0}} \leq M}{\nonumber}
\end{mini!}
and it can be shown that Problem~\eqref{prob:sparse_rec_discretized} is the restriction of~\eqref{prob:sparse_rec_atomic_norm} to the discretized atomic set $\mcA_{d} = \left\{\ba_{\bzeta}\left(\stilde{\theta}\right) \bphi^\transp \left\vert \stilde{\theta} \in \mcV_{\Omega_d}, \bphi \in \mbbC^T \textnormal{ s.t. }\norm{\bphi}{2}=1 \right.\right\} \subset \mcA$. Similarly to~\eqref{prob:sparse_rec_atomic_norm}, given a solution $\bPhi_{\bzeta}\bGamma^\star$ of~\eqref{prob:sparse_rec_discretized}, DoA estimates are given by its support, i.e. $\hat{\btheta} = \supp{\bPhi_{\bzeta}\bGamma^\star}$.

This problem can be approximately solved using SOMP, presented in Algorithm~\ref{algo:SOMP}. At each iteration, SOMP selects the atom of $\bPhi_{\bzeta}$ most correlated with the current residual. The residual is then updated as the component of $\bY$ orthogonal to all atoms selected so far. The angles associated with the selected atoms finally yield the DoA estimates. Note that, in the atom selection step, $\be_t$ denotes the $t$th canonical basis vector of $\mbbC^T$.

\subsection{Limits}\label{subsec:limits}

Both subspace and sparse recovery methods rely on a number of assumptions drawn from~\ref{constraint:narrowband}--\ref{constraint:discrete}. In particular, the theoretical derivations of both families assume narrowband sources, i.e.~\ref{constraint:narrowband}, and their performance is directly governed by the SNR, i.e.~\ref{constraint:snr}. Subspace methods additionally require a sufficient number of snapshots, i.e.~\ref{constraint:snapshots}, for the sample covariance matrix to be an accurate estimate of $\bR_{\bY}$.

Both families also estimate DoAs over the discretized angular domain $\mcV_{\Omega_d}$, i.e.~\ref{constraint:discrete}, so that off-grid sources induce a quantization error, bounded by half the grid step for a uniform grid. Refining the grid reduces this error at the cost of an increased computational complexity and, for sparse recovery methods, of an increased difficulty in discriminating neighboring atoms. Overcoming this trade-off requires decoupling the estimation accuracy from the grid resolution.

Beyond these, subspace methods require non-coherent sources, i.e.~\ref{constraint:non_coh}. Indeed, as the correlation between sources increases, $\bR_{\bS}$ becomes increasingly ill-conditioned, and the separation between the signal and noise subspaces becomes unreliable. In the fully coherent limit, $\bR_{\bS}$ becomes rank deficient, and noise eigenvectors are erroneously assigned to the signal subspace. In that case, the estimated noise subspace is no longer orthogonal to all the steering vectors, which severely degrades the DoA estimation performance.

Finally, both families rely on a perfect knowledge of the array manifold parametrization, i.e.~\ref{constraint:full_knowledge}. Under HWIs, this assumption becomes invalid, directly impacting DoA estimation performance.

Previous work by the authors focused on relaxing~\ref{constraint:full_knowledge} for the MUSIC subspace method~\cite{Chatelier2025}. However, even when considering perfect knowledge of the HWIs, MUSIC performance is still fundamentally limited by the non-coherence assumption~\ref{constraint:non_coh}. Prior work on relaxing this constraint exists, the most notable one being the \textit{spatial smoothing} approach~\cite{Wang94}. However, this approach involves splitting the antenna array into virtual sub-arrays and requires translation-invariant sub-arrays, a condition that is broken in the presence of antenna HWIs. 

This paper therefore proposes to relax~\textnormal{\ref{constraint:full_knowledge}} within the sparse recovery framework, which, unlike subspace methods, remains applicable regardless of the source correlation, even in the fully coherent regime. Additionally, it is also proposed to address the discretization limitation induced by~\textnormal{\ref{constraint:discrete}} by leveraging continuous dictionary learning. Both aspects are detailed in the following section.

\begin{algorithm}[t]
    \caption{$\mathtt{SOMP}(\bY;\bPhi_{\bzeta},K)$}

    \label{algo:SOMP}

    \KwIn{Received signal $\bY$, dictionary $\bPhi_{\bzeta}$, number of \\steps $K$}

    $\bR_0 \gets \bY$ \tcp*{Residual initialization}
    $\mcI \gets \varnothing$ \tcp*{Index set initialization}

    \For{$k \in \llbracket 1,K \rrbracket$}{

        \tcp{Find the most correlated atom}
        $i^{\star}_k \gets \argmax\limits_{i \in \llbracket 1,A \rrbracket} \sum\limits_{t\in \llbracket 1,T \rrbracket} \abs{\langle \bR_{k-1}\be_t, \bphi_{\bzeta,i} \rangle}$\;
        
        \tcp{Update the index set}
        $\mcI \gets \mcI \cup \left\{i^{\star}_k\right\}$\;

        \tcp{Compute the projector (dictionary restricted to selected atoms)}
        $\bP_k \gets \bPhi_{\bzeta,\mcI}\bPhi_{\bzeta,\mcI}^\dagger$\;

        \tcp{Update the residual}
        $\bR_k \gets \bY - \bP_k\bY = \bP_k^{\perp}\bY$\;
    }
    \KwOut{Estimated DoAs $\shat{\btheta} = \left[\stilde{\theta}_i\right]_{i \in \mcI}$, with $\stilde{\theta}_i \in \mcV_{\Omega_d}$}
\end{algorithm}

\section{Proposed method}\label{sec:proposed}

This section presents how the MB-ML paradigm can be leveraged to enhance the performance of sparse recovery methods for joint DoA/HWI estimation.

\subsection{Formal optimization problem}

In order to introduce HWIs estimation and off-grid atoms in~\eqref{prob:sparse_rec_discretized}, i.e. tackling~\ref{constraint:full_knowledge} and~\ref{constraint:discrete}, it is proposed to solve the following optimization problem:
\begin{mini!}
    {\btheta, \bzeta, \bGamma}{\norm{\bY - \bPsi_{\bzeta}\left(\btheta\right)\bGamma}{\mathsf{F}}^2.\tag{$\mathsf{P4}$}\label{problem:proposed_problem}}
    {}{}
    \addConstraint{\bGamma \in \mbbC^{M\times T}}{\nonumber}
    \addConstraint{\btheta \in \mcV_{\Omega}^M}{\nonumber}
    \addConstraint{\bzeta \in \mcZ}{\nonumber}
\end{mini!}
The dictionary $\bPsi_{\bzeta}\left(\btheta\right) = \left[\ba_{\bzeta}\left(\theta_1\right), \cdots, \ba_{\bzeta}\left(\theta_M\right)\right] \in \mbbC^{N \times M}$, hereafter referred to as the \textit{reduced dictionary}, is now a continuous function of DoAs $\btheta \in \mcV_{\Omega}^M$, given the array manifold parametrization $\bzeta \in \mcZ$, restricted to only $M$ atoms. This amounts to assuming the number of sources to be known, an assumption already made by both SOMP and MUSIC. Introducing this a priori knowledge allows dropping the row-sparsity constraint of~\eqref{prob:sparse_rec_discretized}, while restricting its search space.

Remarking that, for fixed $\btheta$ and $\bzeta$, the optimal coefficient matrix in~\eqref{problem:proposed_problem} is given by $\bGamma^\star = \bPsi_{\bzeta}\left(\btheta\right)^\dagger \bY$,~\eqref{problem:proposed_problem} admits the equivalent formulation~\cite{Golub73}:
\begin{mini!}
    {\btheta, \bzeta}{\norm{\bP_{\btheta,\bzeta}^\perp \bY}{\mathsf{F}}^2,\tag{$\mathsf{P5}$}\label{problem:proposed_problem_alternative}}
    {}{}
    \addConstraint{\btheta \in \mcV_{\Omega}^M}{\nonumber}
    \addConstraint{\bzeta \in \mcZ}{\nonumber}
\end{mini!}
where $\bP_{\btheta,\bzeta}^\perp = \mathbf{I}_N - \bPsi_{\bzeta}\left(\btheta\right)\bPsi_{\bzeta}\left(\btheta\right)^\dagger$ defines the projector onto the orthogonal complement of the subspace spanned by the atoms.

\noindent\textbf{Ambiguities.} Let $\mcW = \mbbR^{+,*} \times \mbbR \times \mbbR \times \left[0,2\pi\right)$, and $\bomega = \left(a,b,c,\phi\right) \in \mcW$. Let $\chi_{\bomega}$ denote the transformation mapping all DoAs and array parameters as:
\begin{equation}\label{eq:ambiguity}
    \left\{
    \begin{alignedat}{2}
        &\forall m \in \llbracket 1, M \rrbracket, &\quad& \sin\left(\theta_m\right) \mapsto a \sin\left(\theta_m\right) + b,\\
        &\forall n \in \llbracket 1, N \rrbracket, &\quad& p_n \mapsto \frac{p_n}{a} + c,\\
        &\forall n \in \llbracket 1, N \rrbracket, &\quad& g_n \mapsto \ej[+]{\phi} g_n \ej[+]{\frac{2\pi}{\lambda}\frac{b}{a}p_n},
    \end{alignedat}
    \right.
\end{equation}
and let $\mcG = \left\{\chi_{\bomega}\right\}_{\bomega \in \mcW}$, which can be verified to be a group. The parameters of $\bomega$ respectively correspond to a dilation ($a$) and an offset ($b$) of the DoAs in the sine domain, and a translation ($c$) and global gain phase ($\phi$) of the array.

\begin{proposition}[Ambiguity of~\eqref{problem:proposed_problem_alternative}]\label{prop:ambiguity}
    For every $\bY$, the objective of~\eqref{problem:proposed_problem_alternative} is invariant under $\mcG$, i.e. $\forall \bomega \in \mcW$ such that $\chi_{\bomega}\left(\btheta, \bzeta\right)$ is feasible, $\lVert\bP^\perp_{\chi_{\bomega}\left(\btheta,\bzeta\right)}\bY\rVert_{\mathsf{F}} = \lVert\bP^\perp_{\btheta,\bzeta}\bY\rVert_{\mathsf{F}}$. Thus, if $\left(\shat{\btheta}, \shat{\bzeta}\right)$ is a solution to~\eqref{problem:proposed_problem_alternative}, so is any feasible $\chi_{\bomega}\left(\shat{\btheta}, \shat{\bzeta}\right)$.
\end{proposition}
\begin{proof}
    Let $\left(\btheta', \bzeta'\right) = \chi_{\bomega}\left(\btheta, \bzeta\right)$. Substituting Eq.~\eqref{eq:ambiguity} into Eq.~\eqref{eq:steering_vector} yields, $\forall m \in \llbracket 1, M \rrbracket$, $\ba_{\bzeta'}\left(\theta'_m\right) = \ej[+]{\beta_m}\ba_{\bzeta}\left(\theta_m\right)$, with $\beta_m = \phi - \frac{2\pi}{\lambda}c\sin\left(\theta'_m\right)$. Thus, $\bPsi_{\bzeta'}\left(\btheta'\right) = \bPsi_{\bzeta}\left(\btheta\right)\mathbf{D}$, with $\mathbf{D} = \mathrm{diag}\left(\ej[+]{\beta_1}, \cdots, \ej[+]{\beta_M}\right)$ invertible, so that both matrices span the same subspace. Hence $\bP^\perp_{\btheta',\bzeta'} = \bP^\perp_{\btheta,\bzeta}$.
\end{proof}

Consequently, the minimizers of~\eqref{problem:proposed_problem_alternative} can only be identified up to $\mcG$, and removing this ambiguity requires fixing the four parameters of $\bomega$. The translation $c$ and global phase $\phi$ leave the DoAs unchanged and can for instance be fixed by a reference antenna of known location and gain. In contrast, the dilation $a$ and offset $b$ act on the DoAs through $\sin\left(\theta\right) \mapsto a\sin\left(\theta\right) + b$. Fixing them therefore requires either an assumption on the physical array, e.g. a second reference antenna of known location and gain, or fixed DoAs taking at least two distinct sine values. The latter option, which requires no prior on the array, motivates the sequential approach introduced below.

\subsection{MB-ML approach}\label{subsec:mb_ml}

Following the discussion above, it is proposed to solve~\eqref{problem:proposed_problem_alternative} sequentially using PGD: $\bzeta$ is first learned with fixed DoAs (calibration), which fixes the dilation and offset ambiguities, and is then fixed to estimate $\btheta$ (estimation). As $\bzeta$ is a property of the array rather than of a given measurement, calibration is performed once, offline, whereas estimation runs online for every new measurement. The resulting approach, denoted by UC-C-SOMP, improves SOMP through unsupervised calibration (UC) and continuous dictionaries (C). The receiver is assumed to have access to a calibration dataset $\mcT_{\textrm{cal}} = \left\{\bY_n\right\}_{n=1}^{N_{\textrm{cal}}}$, gathered over time and partitioned into $N_{\mcB}$ batches.\footnote{The impact of $N_{\textrm{cal}}$ on the calibration performance is explored in Section~\ref{sec:experiments}.}

\noindent\textbf{Calibration.} This stage, presented in Algorithm~\ref{algo:UC}, sequentially processes $N_{\mcB}$ batches $\mcB$ of unlabeled measurements $\bY_i$, whose underlying DoAs $\btheta_i$ are unknown. For each batch, the estimated DoAs $\shat{\btheta}_i$ are obtained by SOMP with the dictionary $\bPhi_{\bzeta_0} \in \mbbC^{N \times A}$, built on $\mcV_{\Omega_d}$ from the nominal array parametrization for the first batch and from the previous batch estimate otherwise. These DoAs are then frozen, and $\bzeta$ is learned by PGD on the objective of~\eqref{problem:proposed_problem_alternative}, evaluated with the reduced dictionaries $\bPsi_{\bzeta}\left(\shat{\btheta}_i\right) \in \mbbC^{N \times M}$ and averaged over the batch, each gradient step being followed by the projection:
\begin{equation}\label{eq:proj_zeta}
    \bPi_{\bzeta}: \begin{cases}
        p_n \mapsto \tilde{p}_n + \max\left(-\frac{\lambda}{4}, \min\left(p_n - \tilde{p}_n, \frac{\lambda}{4}\right)\right),\\
        \bg \mapsto \frac{\bg}{\norm{\bg}{2}},
    \end{cases}
\end{equation}
where $\tilde{p}_n$ denotes the nominal location of the $n$th antenna. $\bPi_{\bzeta}$ restricts the search to physically plausible arrays: antennas remain within $\lambda/4$ of their nominal locations, so that they cannot cross, and the gains are normalized consistently with Eq.~\eqref{eq:steering_vector}. The learned parametrization $\hat{\bzeta}$ initializes the PGD of the next batch.

\begin{remark}[Residual ambiguity]\label{remark:residual_ambiguity}
    Freezing the DoAs fixes the dilation $a$ and offset $b$ of Proposition~\ref{prop:ambiguity}. However, as these DoAs are estimated with the nominal array, the learned array and the subsequent DoA estimates remain subject to a global dilation and offset with respect to the physical ones, which is expected to only marginally degrade the estimation performance for realistic HWIs. The array parameters are further only identified up to the translation $c$ and global phase $\phi$, which do not affect the DoAs. Removing these residual ambiguities would require prior knowledge, e.g. the locations and gains of two reference antennas, which the proposed method deliberately avoids.
\end{remark}

\noindent\textbf{Estimation.} This stage, presented in Algorithm~\ref{algo:C_SOMP}, is applied online to each new measurement $\bY$. First, SOMP, equipped with the dictionary $\bPhi_{\shat{\bzeta}}$ rebuilt from the previously learned array parametrization, provides initial DoA estimates. These are then refined by PGD on the objective of~\eqref{problem:proposed_problem_alternative} with respect to $\btheta$ only, each gradient step being followed by the projection $\bPi_{\btheta}: \theta_m \mapsto \arcsin\left(\sin\left(\theta_m\right)\right)$, which ensures that estimated DoAs belong to $\mcV_{\Omega}$. Thanks to this continuous refinement, the DoA estimates are no longer confined to the discrete angular domain $\mcV_{\Omega_d}$, which addresses~\ref{constraint:discrete}.

\begin{remark}[Variants and naming]\label{remark:variants}
    The proposed calibration is unsupervised, as it requires no labeled measurement-DoA pairs. A supervised variant, SC-C-SOMP, uses the true DoAs in place of the estimated ones during calibration, which fixes the dilation and offset to their physical values. More generally, the first prefix denotes the calibration strategy, i.e. UC, SC or none, and the second one the dictionary over which the DoAs are estimated, i.e. continuous (C) or discrete (D). As calibration and estimation are independent, every combination is valid, e.g. D-SOMP denotes the classical SOMP algorithm, C-SOMP its continuous counterpart without calibration, and UC-D-SOMP and SC-D-SOMP the calibrated variants over a discrete dictionary.
\end{remark}

\begin{algorithm}[t]
    \caption{$\mathtt{UC}\left(\mcB;\bzeta_0,N_{\nabla_{\bzeta}},M,\mcV_{\Omega_d}\right)$}

    \label{algo:UC}

    \KwIn{Calibration batch $\mcB$, initial array param. $\bzeta_0$,\\ number of PGD steps $N_{\nabla_{\bzeta}}$, number of sources $M$,\\ discretized angular domain $\mcV_{\Omega_d}$.}

    \tcc{Construct dictionary based on\\ initial array}
    $\bPhi_{\bzeta_0} \gets \left[\ba_{\bzeta_0}\left(\stilde{\theta}_1\right), \cdots, \ba_{\bzeta_0}\left(\stilde{\theta}_A\right)\right],\ \stilde{\theta}_i \in \mcV_{\Omega_d}$\;

    \tcc{Estimate DoAs with initial array}
    $\shat{\btheta}_n \gets \mathtt{SOMP}\left(\bY_n; \bPhi_{\bzeta_0}, M\right),\ \forall \bY_n \in \mcB$\;

    \tcp{Initialize}
    $\bzeta^{\left(0\right)} \gets \bzeta_0$\;

    \For{$k \in \llbracket 0,N_{\nabla_{\bzeta}}-1 \rrbracket$}{

        \For{$n \in \llbracket 1, \abs{\mcB} \rrbracket$}{
            \tcp{Update reduced dictionary}
            $\bPsi_{\bzeta^{\left(k\right)}}\left(\shat{\btheta}_n\right) \gets \left[\ba_{\bzeta^{\left(k\right)}} \left(\shat{\theta}_{n,1}\right), \cdots, \ba_{\bzeta^{\left(k\right)}} \left(\shat{\theta}_{n,M}\right)\right]$\;

            \tcp{Compute the projector}
            $\bP^\perp_{\shat{\btheta}_n, \bzeta^{\left(k\right)}} \gets \mathbf{I}_N - \bPsi_{\bzeta^{\left(k\right)}}\left(\shat{\btheta}_n\right)\bPsi_{\bzeta^{\left(k\right)}}\left(\shat{\btheta}_n\right)^\dagger$\;
        }

        \tcp{Compute the batch loss}
        $\mcJ_{\mcB}\left(\bzeta^{\left(k\right)}\right) \gets \dfrac{1}{\abs{\mcB}}\sum\limits_{n \in \llbracket 1, \abs{\mcB} \rrbracket} \norm{\bP^\perp_{\shat{\btheta}_n,\bzeta^{\left(k\right)}}\bY_n}{\mathsf{F}}^2$\;

        \tcp{Update array parametrization}
        $\bzeta^{\left(k+1\right)} \gets \bzeta^{\left(k\right)} - \alpha_{\bzeta} \nabla_{\bzeta^{\left(k\right)}} \mcJ_{\mcB}\left(\bzeta^{\left(k\right)}\right)$\;

        \tcp{Project on the feasible set}
        $\bzeta^{\left(k+1\right)} \gets \bPi_{\bzeta}\left( \bzeta^{\left(k+1\right)}\right)$\;
    }
    \tcp{Finalize calibration}
    $\shat{\bzeta} \gets \bzeta^{\left(N_{\nabla_{\bzeta}}\right)}$\;
    \KwOut{Estimated array parametrization $\shat{\bzeta}$}
\end{algorithm}

\begin{algorithm}[t]
    \caption{$\mathtt{C}$-$\mathtt{SOMP}\left(\bY;\bzeta,N_{\nabla_{\btheta}},M,\mcV_{\Omega_d}\right)$}

    \label{algo:C_SOMP}

    \KwIn{Received signal $\bY$, array parametrization $\bzeta$,\\ number of PGD steps $N_{\nabla_{\btheta}}$, number of sources $M$,\\ discretized angular domain $\mcV_{\Omega_d}$.}

    \tcc{Construct discrete dictionary}
    $\bPhi_{\bzeta} \gets \left[\ba_{\bzeta}\left(\stilde{\theta}_1\right), \cdots, \ba_{\bzeta}\left(\stilde{\theta}_A\right)\right],\ \stilde{\theta}_i \in \mcV_{\Omega_d}$\;

    \tcc{Initialize DoAs}
    $\btheta^{\left(0\right)} \gets \mathtt{SOMP}\left(\bY; \bPhi_{\bzeta}, M\right)$\;

    \For{$k \in \llbracket 0,N_{\nabla_{\btheta}}-1 \rrbracket$}{

        \tcp{Update reduced dictionary}
        $\bPsi_{\bzeta}\left(\btheta^{\left(k\right)}\right) \gets \left[\ba_{\bzeta} \left(\theta^{\left(k\right)}_1\right), \cdots, \ba_{\bzeta} \left(\theta^{\left(k\right)}_M\right)\right]$\;

        \tcp{Compute the projector}
        $\bP^\perp_{\btheta^{\left(k\right)}, \bzeta} \gets \mathbf{I}_N - \bPsi_{\bzeta}\left(\btheta^{\left(k\right)}\right)\bPsi_{\bzeta}\left(\btheta^{\left(k\right)}\right)^\dagger$\;

        \tcp{Compute the loss}
        $\mcJ\left(\btheta^{\left(k\right)}\right) \gets \norm{\bP^\perp_{\btheta^{\left(k\right)},\bzeta}\bY}{\mathsf{F}}^2$\;

        \tcp{Update estimated DoAs}
        $\btheta^{\left(k+1\right)} \gets \btheta^{\left(k\right)} - \alpha_{\btheta} \nabla_{\btheta^{\left(k\right)}} \mcJ\left(\btheta^{\left(k\right)}\right)$\;

        \tcp{Project on the feasible set}
        $\btheta^{\left(k+1\right)} \gets \bPi_{\btheta} \left(\btheta^{\left(k+1\right)}\right)$\;
    }

    \tcp{Finalize estimation}
    $\shat{\btheta} \gets \btheta^{\left(N_{\nabla_{\btheta}}\right)}$\;
    \KwOut{Estimated DoAs $\shat{\btheta}$}
\end{algorithm}

\subsection{Complexity analysis}\label{subsec:complexity}

It is proposed to derive the computational complexity of the calibration and estimation methods, and to compare it with that of the MUSIC and D-SOMP baselines, an overview being given in Table~\ref{tab:complexity}.

\noindent\textbf{Baselines.} Both baselines are considered over the same discretized angular domain $\mcV_{\Omega_d}$, with $\abs{\mcV_{\Omega_d}} = A$. The MUSIC algorithm involves (i) computing the sample covariance matrix, which costs $\mcO\left(N^2 T\right)$, (ii) performing its EVD, which costs $\mcO\left(N^3\right)$, and (iii) evaluating the MUSIC spectrum on $\mcV_{\Omega_d}$, which costs $\mcO\left(AN\left(N-M\right)\right)$, resulting in an overall $\mcO\left(N^2 T + N^3 + AN\left(N-M\right)\right)$ complexity. The D-SOMP algorithm performs $M$ iterations, each involving (i) correlating the dictionary with the current residual, which costs $\mcO\left(ANT\right)$, (ii) selecting the atom maximizing the row norm of the resulting correlation matrix, which costs $\mcO\left(AT\right)$, and (iii) updating the least-squares coefficients and the residual, which is bounded by $\mcO\left(MNT + M^2N + M^3\right)$. As $M \ll A$, the correlation step dominates, yielding an overall $\mcO\left(AMNT\right)$ complexity.

\noindent\textbf{Calibration.} The proposed unsupervised calibration method involves three main steps: (i) DoA initialization using SOMP with a full dictionary $\bPhi_{\bzeta_0} \in \mbbC^{N \times A}$, performed on each element of the batch, which costs $\mcO\left(\abs{\mcB}AMNT\right)$, (ii) building the updated reduced dictionaries $\bPsi_{\bzeta^{\left(k\right)}}\left(\shat{\btheta}_n\right)$, which costs $\mcO\left(MN\right)$ for each batch element and is absorbed in the previous term, and (iii) $N_{\nabla_{\bzeta}}$ PGD iterations, each evaluating the batch loss over the $\abs{\mcB}$ measurements at a cost of $\mcO\left(MNT+M^2N+M^3\right)$ per measurement. As the backward pass of automatic differentiation costs a constant multiple of the forward one, irrespective of the number of optimized parameters, and as the projection $\bPi_{\bzeta}$ is performed element-wise in $\mcO\left(N\right)$, the per-batch calibration complexity is $\mcO\left(\abs{\mcB}\left(AMNT + N_{\nabla_{\bzeta}}\left(MNT + M^2N + M^3\right)\right)\right)$. Since the algorithm is applied to the $N_{\mcB}$ batches partitioning $\mcT_{\textrm{cal}}$, with $N_{\mcB}\abs{\mcB} = N_{\textrm{cal}}$ and considering $M \ll A$  and $M \ll T$, the overall calibration complexity reduces to $\mcO\left(N_{\textrm{cal}}\left(A + N_{\nabla_{\bzeta}}\right)MNT\right)$.

\noindent\textbf{Estimation.} The estimation method follows the same three-step structure as the calibration approach, $\btheta$ being now the optimization variable. It is applied to a single measurement, thereby costing $\mcO\left(\left(A+N_{\nabla_{\btheta}}\right) MNT\right)$ when considering $M \ll A$  and $M \ll T$.

\begin{remark}[Cost comparison]\label{remark:cost}
    Both the unsupervised calibration and estimation stages are dominated by their SOMP initialization as soon as $N_{\nabla_{\bzeta}}, N_{\nabla_{\btheta}} \ll A$. The UC calibration stage is roughly $N_{\mathrm{cal}}$ times more costly than a single D-SOMP run, but is performed only once, so that its cost amortizes over all subsequent measurements. The SC variant, relying on known DoAs, skips the SOMP initialization, so that its calibration cost no longer depends on the grid size $A$. During inference, the continuous refinement decouples the estimation precision from the dictionary resolution, so that a given accuracy can be reached with a dictionary of size $A_{\mathtt{C}}$, much smaller than the size $A_{\mathtt{D}}$ required by D-SOMP. At equal estimation accuracy, C-SOMP is therefore less complex than D-SOMP whenever $A_{\mathtt{C}} + \kappa N_{\nabla_{\btheta}} < A_{\mathtt{D}}$, where $\kappa$ denotes the cost ratio between a PGD iteration and a single atom correlation. This condition is assessed empirically in Fig.~\ref{fig:pareto_complexity}. MUSIC remains the least complex method, its per-atom cost $N\left(N-M\right)$ being much lower than the $MNT$ of SOMP-based methods, but it breaks down in the presence of coherent sources and still relies on a discrete angular grid.
\end{remark}

\begin{table*}[t]
    \centering
    \setlength{\tabcolsep}{4pt}\renewcommand{\arraystretch}{1.3}
    \caption{Computational complexity comparison, given $M \ll A$ and $M \ll T$.}
    \label{tab:complexity}
    \begin{tabular}{lccccc}
        \toprule
        & \multicolumn{2}{c}{\textbf{Baselines}} & \multicolumn{2}{c}{\textbf{Calibration}} & \textbf{Estimation} \\
        \cmidrule(lr){2-3} \cmidrule(lr){4-5} \cmidrule(lr){6-6}
        \textbf{Method} & MUSIC & D-SOMP & UC & SC & C-SOMP \\
        \midrule
        \textbf{Complexity} & $\mcO\left(N^2T + N^3 + AN\left(N-M\right)\right)$ & $\mcO\left(AMNT\right)$ & $\mcO\left(N_{\textrm{cal}}\left(A + N_{\nabla_{\bzeta}}\right)MNT\right)$ & $\mcO\left(N_{\textrm{cal}} N_{\nabla_{\bzeta}} M N T\right)$ & $\mcO\left(\left(A + N_{\nabla_{\btheta}}\right) M N T\right)$ \\
        \bottomrule
    \end{tabular}
\end{table*}

\subsection{Conceptual comparison with differentiable subspace methods}

\begin{figure}[tb]
    \centering
    \includegraphics[width=.95\columnwidth]{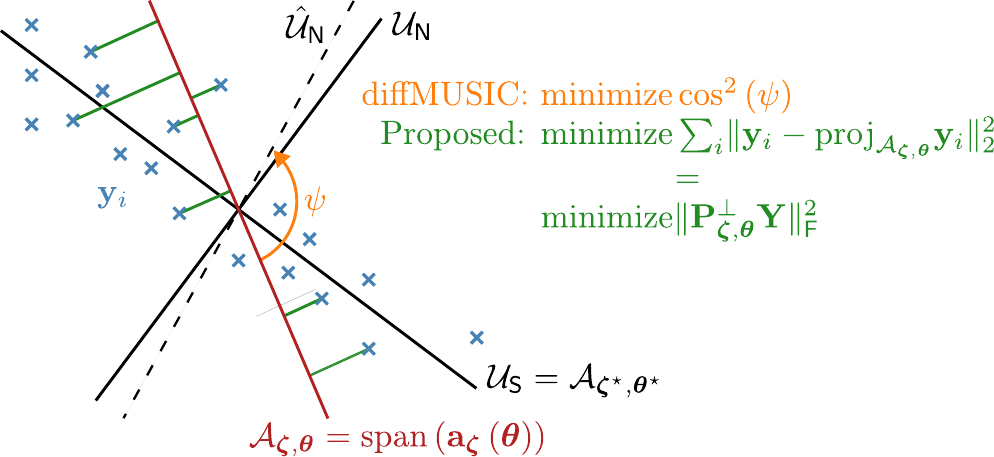}
    \caption{High-level comparison between diffMUSIC~\cite{Chatelier2025} and the proposed approach, for $N=2$ and a single source. The angle $\psi$ denotes the principal angle between the current signal $\mcA_{\bzeta,\btheta}$ and noise $\hat{\mcU}_{\mathsf{N}}$ subspaces. As the number of snapshots grows $\hat{\mcU}_{\mathsf{N}} \rightarrow \mcU_{\mathsf{N}}$.}
    \label{fig:subspace}
\end{figure}

Fig.~\ref{fig:subspace} illustrates the conceptual differences between the proposed method, i.e. solving~\eqref{problem:proposed_problem_alternative}, and the differentiable MUSIC (diffMUSIC) of~\cite{Chatelier2025}, in the presence of a single source. $\mcA_{\bzeta,\btheta}$ denotes the current signal subspace, which suffers from HWIs, whereas $\mcU_{\mathsf{S}}$ and $\mcU_{\mathsf{N}}$ respectively denote the signal and noise subspaces, $\mcU_{\mathsf{S}}$ coinciding with the subspace $\mcA_{\bzeta^\star,\btheta^\star}$ associated with the true array parametrization and DoAs. From a high level perspective, both the proposed method and diffMUSIC aim at aligning the current subspace $\mcA_{\bzeta,\btheta}$ with $\mcU_{\mathsf{S}}$. However, they rely on fundamentally different criteria to do so, as detailed below.

\begin{proposition}[Chordal-distance interpretation of diffMUSIC]\label{prop:diffMUSIC_principal_angles}
    Consider leveraging diffMUSIC~\cite{Chatelier2025} to minimize the inverse MUSIC spectrum objective:
    \begin{equation}\label{eq:mu_sl_p}
        \mcJ_{\Theta}\left(\bzeta\right) = \frac{1}{\abs{\mcT}}\sum_{\left(\btheta,\bY\right) \in \mcT} \norm{\hat{\bU}_{\mathsf{N}}^\herm \bA_{\bzeta}\left(\btheta\right)}{\mathsf{F}}^2,
    \end{equation}
    where $\hat{\bU}_{\mathsf{N}}$ is estimated from $\bY$. For full column rank $\bA_{\bzeta}\left(\btheta\right)$ and $M \leq N-M$, one has:
    \begin{equation}\label{eq:chordal_bounds}
        \sigma_{\min}^2\left(M - d_c^2\right) \leq \norm{\hat{\bU}_{\mathsf{N}}^\herm \bA_{\bzeta}\left(\btheta\right)}{\mathsf{F}}^2 \leq \sigma_{\max}^2\left(M - d_c^2\right),
    \end{equation}
    where $d_c = d_c\left(\mcA_{\bzeta,\btheta},\shat{\mcU}_{\mathsf{N}}\right)$ is the chordal distance between the current signal subspace $\mcA_{\bzeta,\btheta}$ and the noise subspace $\hat{\mcU}_{\mathsf{N}}$. The extreme singular values $\sigma_{\min}$ and $\sigma_{\max}$ of $\bA_{\bzeta}\left(\btheta\right)$ account for the non-orthogonality of the steering vectors: if the latter are orthogonal, both equal $1$ and Eq.~\eqref{eq:chordal_bounds} holds with equality. Minimizing $\mcJ_{\Theta}\left(\bzeta\right)$ thus amounts to maximizing the chordal distance between $\mcA_{\bzeta,\btheta}$ and $\hat{\mcU}_{\mathsf{N}}$, i.e. orthogonalizing them in Fig.~\ref{fig:subspace}.
\end{proposition}
\begin{proof}
    See Appendix~\ref{app:proof_prop_diffMUSIC}.  
\end{proof}

As presented in Proposition~\ref{prop:diffMUSIC_principal_angles}, diffMUSIC seeks orthogonality between $\mcA_{\bzeta,\btheta}$ and $\hat{\mcU}_{\mathsf{N}}$ through their chordal distance, which, in Fig.~\ref{fig:subspace}, amounts to minimizing $\cos^2\left(\psi\right)$. This criterion is purely angular: $\bY$ only intervenes through the noise subspace estimated from $\hat{\bR}_{\bY}$. Its relevance is therefore entirely conditioned on the quality of the subspace separation, which, as discussed in Section~\ref{sec:classical}, requires both a sufficient number of snapshots and non-coherent sources.

The proposed method, on the contrary, measures the residual energy left by projecting the snapshots onto $\mcA_{\bzeta,\btheta}$, depicted by the green segments in Fig.~\ref{fig:subspace}. It thus directly quantifies how well the raw measurements are explained by the current array manifold knowledge, without resorting to any covariance estimation nor subspace separation. Consequently, its validity does not depend on the source correlation, which makes it applicable in the coherent regime.

\section{Experiments}\label{sec:experiments}

This section presents the evaluation framework, imperfection models, baselines and experimental results.

\subsection{Evaluation framework}

\noindent\textbf{Dataset generation.} As in~\cite{Chatelier2025}, a ULA with $N$ antennas and $\lambda/2$ inter-element spacing is considered. The operating frequency is set to \qty{2.4}{\GHz}, and measurements are generated following Eq.~\eqref{eq:received_signal_matrix}. The source DoAs are drawn as:
\begin{equation}
    \forall i \in \llbracket 1,M \rrbracket,\ \theta_i \sim \mcU\left(\left[-80^\circ,80^\circ\right]\right).
\end{equation}

\begin{remark}[Regularized objective for SC]
    For the supervised calibration method, two very similar DoAs make the reduced dictionary ill-conditioned, so that its pseudo-inverse is numerically unstable, which was observed to make the calibration loss diverge. The Tikhonov-regularized variant of~\eqref{problem:proposed_problem}, $\norm{\bY - \bPsi_{\bzeta}\left(\btheta\right)\bGamma}{\mathsf{F}}^2 + \mu \norm{\bGamma}{\mathsf{F}}^2$ with $\mu > 0$, is therefore used, which replaces the objective of~\eqref{problem:proposed_problem_alternative} by $\mathsf{Tr}(\bY^\herm \bP^{\perp}_{\btheta,\bzeta,\mu} \bY)$, with $\bP^{\perp}_{\btheta,\bzeta,\mu} = \bI_N - \bPsi_{\bzeta}(\btheta)(\bPsi^\herm_{\bzeta}(\btheta) \bPsi_{\bzeta}(\btheta) + \mu \bI_M)^{-1} \bPsi^\herm_{\bzeta}(\btheta)$. The inverse is well defined for any DoA separation, and $\bP^{\perp}_{\btheta,\bzeta,\mu} \rightarrow \bP^{\perp}_{\btheta,\bzeta}$ as $\mu \rightarrow 0$. In the following, $\mu = 10^{-4}$.
\end{remark}

Two source signal configurations are considered: (i) non-coherent sources, i.e. $\forall i \in \llbracket 1, T \rrbracket,\ \bs_i \sim \mcCN\left(\mathbf{0}_M,\sigma^2_s \mathbf{I}_M\right)$, and (ii) fully-coherent sources, where every source emits the same signal, i.e. $\bS = \mathbf{1}_M \bu^\transp$ with $\bu \in \mbbC^T \sim \mcCN\left(\mathbf{0}_T,\sigma^2_s \mathbf{I}_T\right)$, which violates~\ref{constraint:non_coh}. The sensing SNR is defined as $\text{SNR} = 10 \log_{10}\left(\sigma^2_s/\sigma^2_n\right)$.

Both antenna locations and gain HWIs are considered. In Eq.~\eqref{eq:steering_vector}, one has~\cite{Chatelier2025,Shmuel2025,mateosramos2026}:
\begin{equation}
    \forall i \in \llbracket 1, N \rrbracket,\ p_i = \tilde{p}_i + \delta_{p_i},
\end{equation}
where $p_i$ and $\tilde{p}_i$ respectively denote the \textit{physical} and \textit{nominal} antenna locations, $\delta_{p_i} \sim \mcU\left(\left[-\eta,\eta\right]\right)$ representing the HWI-induced deviation from the nominal location, with $\eta < \lambda/4$ so that antennas cannot cross. Similarly, one has~\cite{Yassine22,Chatelier2025}:
\begin{equation}
    \forall i \in \llbracket 1, N \rrbracket,\ g_i = \tilde{g}_i + \delta_{g_i},
\end{equation}
with $g_i$ and $\tilde{g}_i$ respectively denoting the \textit{physical} and \textit{nominal} antenna gains, with $\tilde{g}_i = 1,\ \forall i \in \llbracket 1, N \rrbracket$, and $\delta_{g_i} \sim \mcCN\left(0,\sigma^2_{g}\right)$ representing the HWI perturbation.

\noindent\textbf{Simulation parameters.} Unless explicitly mentioned, $N=16$ antennas, $M=5$ sources, $T=100$ snapshots, $\eta = \lambda/8$ and $\sigma^2_{g} = 0.1$ are considered in the following simulations. Additionally, the calibration dataset is composed of $N_{\textrm{cal}} = 500$ samples, split into $N_{\mcB} = 5$ batches of $\abs{\mcB} = 100$ samples. Calibration and estimation use $N_{\nabla_{\bzeta}} = 200$ and $N_{\nabla_{\btheta}} = 100$ Adam steps, with learning rates $\alpha_{\bzeta} = 10^{-3}$ and $\alpha_{\btheta} = 10^{-2}$, respectively. For each method requiring it, the discrete angular grid $\mcV_{\Omega_d}$ is obtained through a uniform sampling of $\mcV_{\Omega}$ with $A=50$ points.\footnote{Note that, given prior knowledge of the problem at hand, non-uniform sampling strategies can be envisioned, but remain beyond the scope of this paper.}

\noindent\textbf{Baselines.} Both MUSIC and SOMP baselines are considered. Given that their performance depends on knowledge of the array manifold parametrization, both nominal and physical variants, i.e. imperfect and perfect knowledge, will be considered.

\noindent\textbf{Evaluation metrics.} As in~\cite{Shmuel2025,Chatelier2025,Gast2026}, estimation performance is measured by the root mean squared periodic error (RMSPE), which accounts for both the periodicity of angles and the permutation invariance of the estimates. Given a test set of measurements and associated DoAs $\mcT_{\textrm{test}} = \left\{\left(\bY_n, \btheta_n\right)\right\}_{n=1}^{N_{\textrm{test}}}$, the RMSPE is defined as:
\begin{equation}
    \mu_{\btheta} = \dfrac{1}{N_{\textrm{test}}} \sum_{\left(\btheta,\bY\right)\in \mcT_{\mathrm{test}}} \min_{\bP \in \mcP_M}\frac{\norm{\mathrm{mod}_{\pi}\left(\btheta - \bP\hat{\btheta}\left(\bY\right)\right)}{2}}{\sqrt{M}},
\end{equation}
where $\hat{\btheta}\left(\bY\right)$ is the estimated DoA for measurement $\bY$, and $\mcP_M$ denotes the set of $M\times M$ permutation matrices. For all experiments, performance is evaluated on $N_{\textrm{test}} = 1000$ different DoA vectors and measurements. Note that, in the following results, the median RMSPE over $100$ HWI realizations is reported.

\subsection{Experimental results}

\begin{figure*}[htbp]
    \centering
    \begin{minipage}[b]{\columnwidth}
        \centering
        \includegraphics[width=.9\columnwidth]{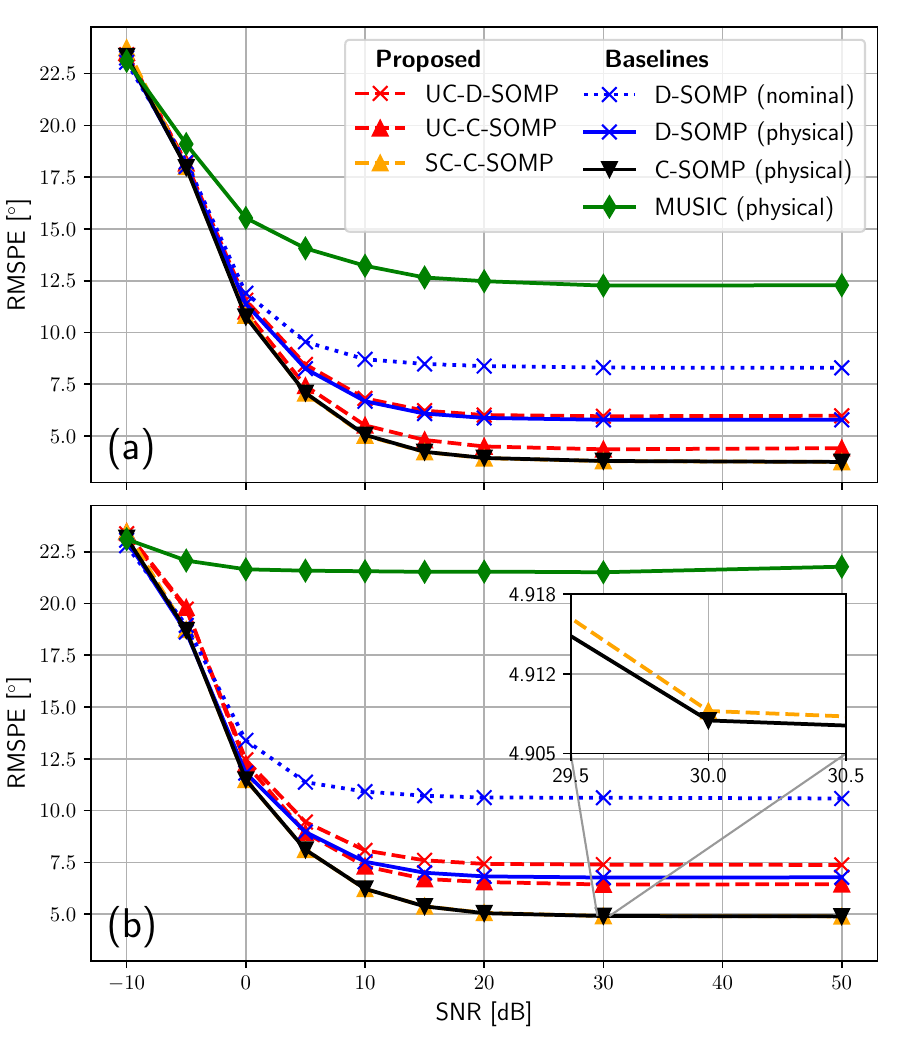}
        \caption{Performance against varying noise levels: ($\textsf{a}$) non-coherent signals, ($\textsf{b}$) coherent signals. ($M=5$, $N=16$, $T=100$, $A=50$).}
        \label{fig:multi_snr}
    \end{minipage}
    \hfill
    \begin{minipage}[b]{\columnwidth}
        \centering
        \includegraphics[width=.9\columnwidth]{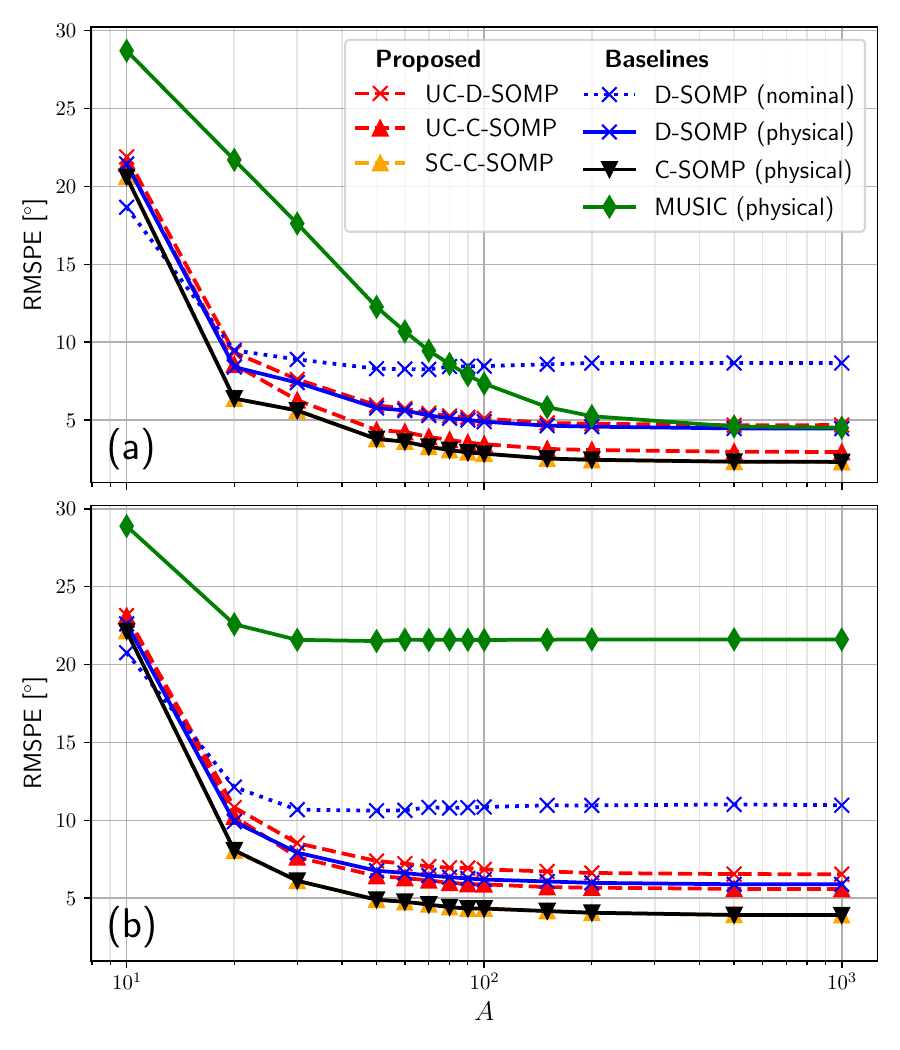}
        \caption{Performance against varying $A$: ($\textsf{a}$) non-coherent signals, ($\textsf{b}$) coherent signals. ($M=5$, $N=16$, $T=100$, \qty{30}{dB} SNR).}
        \label{fig:multi_atoms}
    \end{minipage}
\end{figure*}

\noindent\textbf{Performance against sensing noise.} Fig.~\ref{fig:multi_snr} reports the performance of the proposed methods as a function of the SNR, with $A=50$, for both non-coherent and coherent sources, thereby evaluating the impact of~\ref{constraint:non_coh} and~\ref{constraint:snr}. D-SOMP with nominal knowledge, i.e. without knowledge of the HWIs, constitutes the classical baseline, while C-SOMP with physical knowledge, i.e. with perfect knowledge of the HWIs, acts as an oracle bound. First, all sparse recovery methods outperform MUSIC, even when the latter is equipped with physical knowledge, and, as expected, remain effective in the fully coherent setting. Second, both calibration strategies are effective: SC-C-SOMP is indistinguishable from the oracle C-SOMP, which holds in all subsequent results, and UC-D-SOMP closely approaches D-SOMP with physical knowledge. More importantly, the fully unsupervised UC-C-SOMP outperforms all discrete baselines, including D-SOMP with physical knowledge.

\noindent\textbf{Performance against angular grid precision.} Fig.~\ref{fig:multi_atoms} presents the performance of all methods as a function of $A$, for a fixed SNR of \qty{30}{dB}. As the angular resolution of $\mcV_{\Omega_d}$ improves with $A$, this experiment evaluates the impact of~\ref{constraint:discrete}. As expected, all methods degrade with coarse grids. Nevertheless, the continuous-dictionary variants, i.e. UC-C-SOMP, SC-C-SOMP and C-SOMP, attain a performance level that the discrete alternatives do not reach, even with very fine grids. In particular, with $A = 50$, the continuous variants already match or outperform their discrete counterparts with $A = 10^3$, confirming that a much coarser dictionary is sufficient, as anticipated in Remark~\ref{remark:cost}. Finally, MUSIC converges to the performance of D-SOMP as $A$ increases, although only in the non-coherent setting.

\begin{figure}[t]
    \centering
    \includegraphics[width=.9\columnwidth]{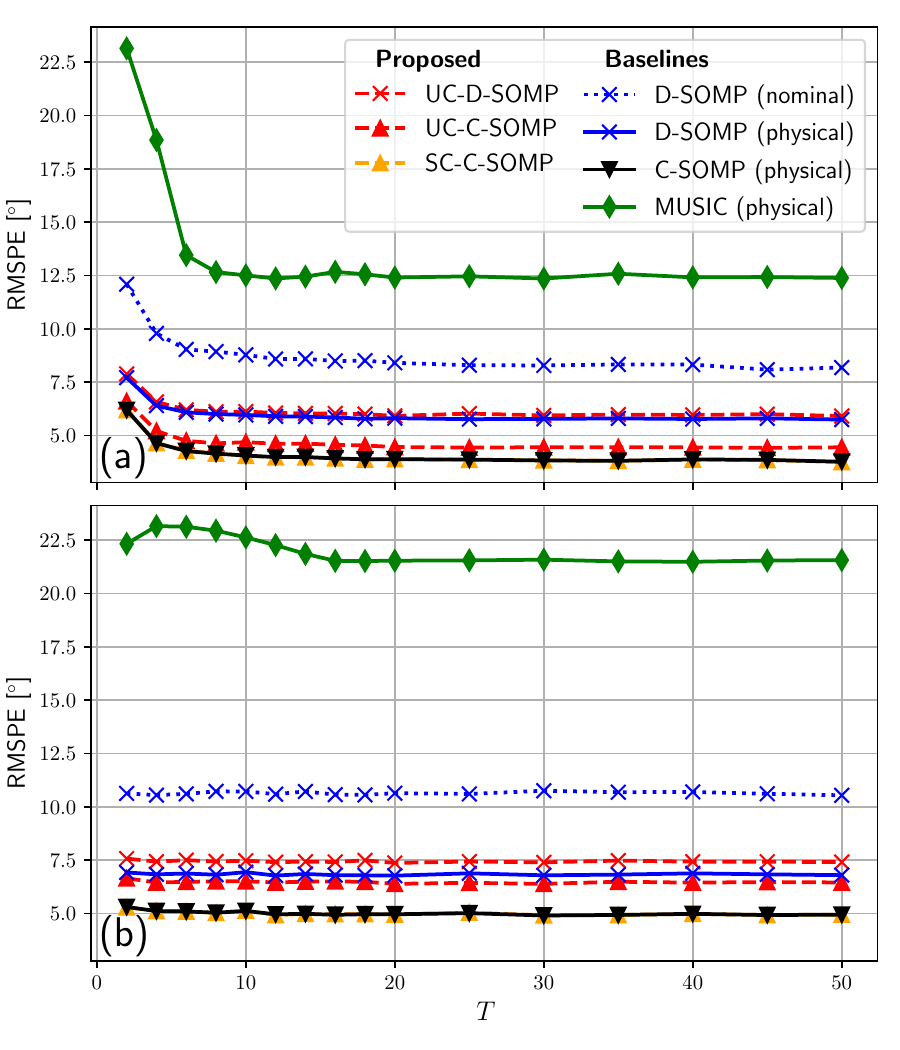}
    \caption{Performance against varying $T$: ($\textsf{a}$) non-coherent signals, ($\textsf{b}$) coherent signals. ($M=5$, $N=16$, $A=50$, \qty{30}{dB} SNR).}
    \label{fig:multi_T}
\end{figure}

\noindent\textbf{Performance against snapshots number.} Fig.~\ref{fig:multi_T} presents the performance for different numbers of snapshots, thereby evaluating the impact of~\ref{constraint:snapshots}, for a fixed SNR of \qty{30}{dB} and $A=50$ angular points. As expected, the performance of MUSIC improves with $T$ in the non-coherent setting, which directly follows from a more accurate sample covariance matrix, translating into an improved subspace separation. This is not the case in the coherent setting, where the source covariance remains rank-deficient irrespective of $T$. In contrast, the performance of the proposed methods depends only weakly on $T$: it saturates from $T \approx 6$ in the non-coherent setting and remains virtually unaffected by $T$ in the coherent setting. Since joint calibration and estimation from few snapshots is a notably challenging problem, this robustness constitutes a significant practical advantage.

\begin{figure}[h]
    \centering
    \includegraphics[width=.9\columnwidth]{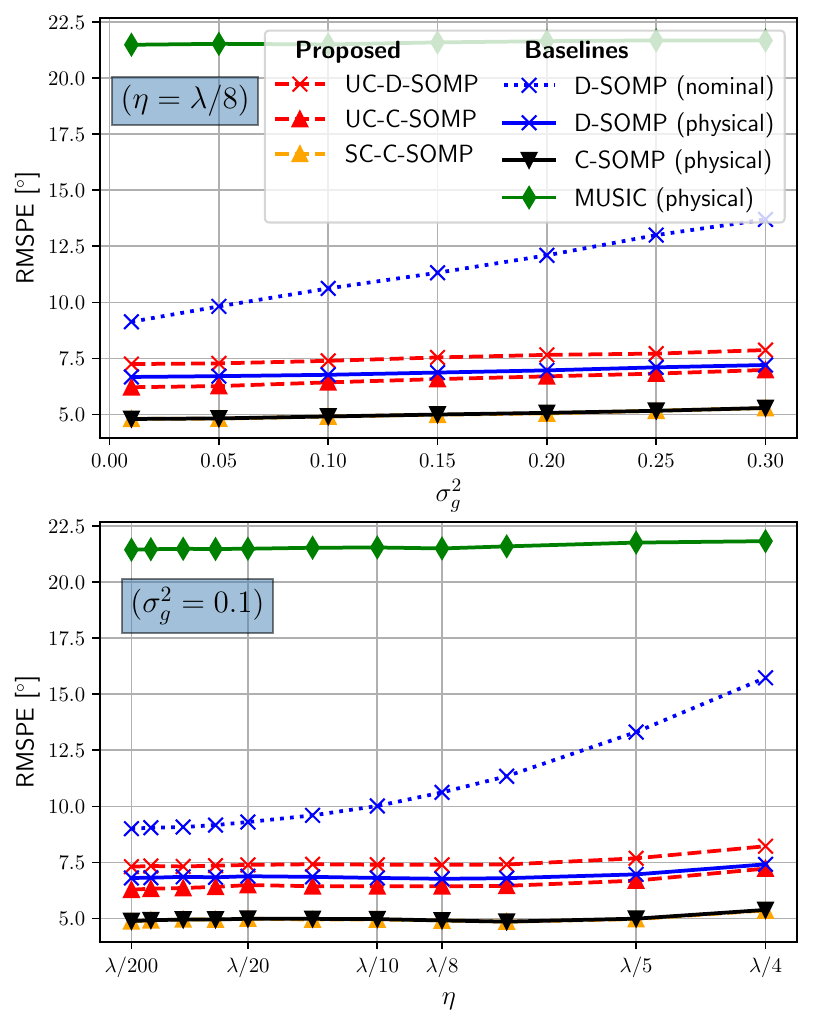}
    \caption{Performance against varying $\sigma^2_g$ and $\eta$ values, for coherent signals. ($M=5$, $N=16$, $T=100$, $A=50$, \qty{30}{dB} SNR).}
    \label{fig:hwi_strength}
\end{figure}

\noindent\textbf{Performance against varying HWI strength.} Fig.~\ref{fig:hwi_strength} presents the performance obtained with increasing HWI severity, i.e. with growing gain noise variance $\sigma^2_g$ and maximum antenna displacement $\eta$, in the coherent setting. As expected, the performance of D-SOMP with nominal array knowledge degrades as the HWI strength increases, for both impairment types. In contrast, the performance of the calibrated variants remains essentially constant over the whole range, each of them reaching that of their counterparts with physical array knowledge. The proposed calibration strategies are therefore insensitive to the impairment severity, even in the degenerate regime where the array geometry approaches element overlap, i.e. $\eta \rightarrow \lambda/4 $.

\begin{figure}[t]
    \centering
    \includegraphics[width=.9\columnwidth]{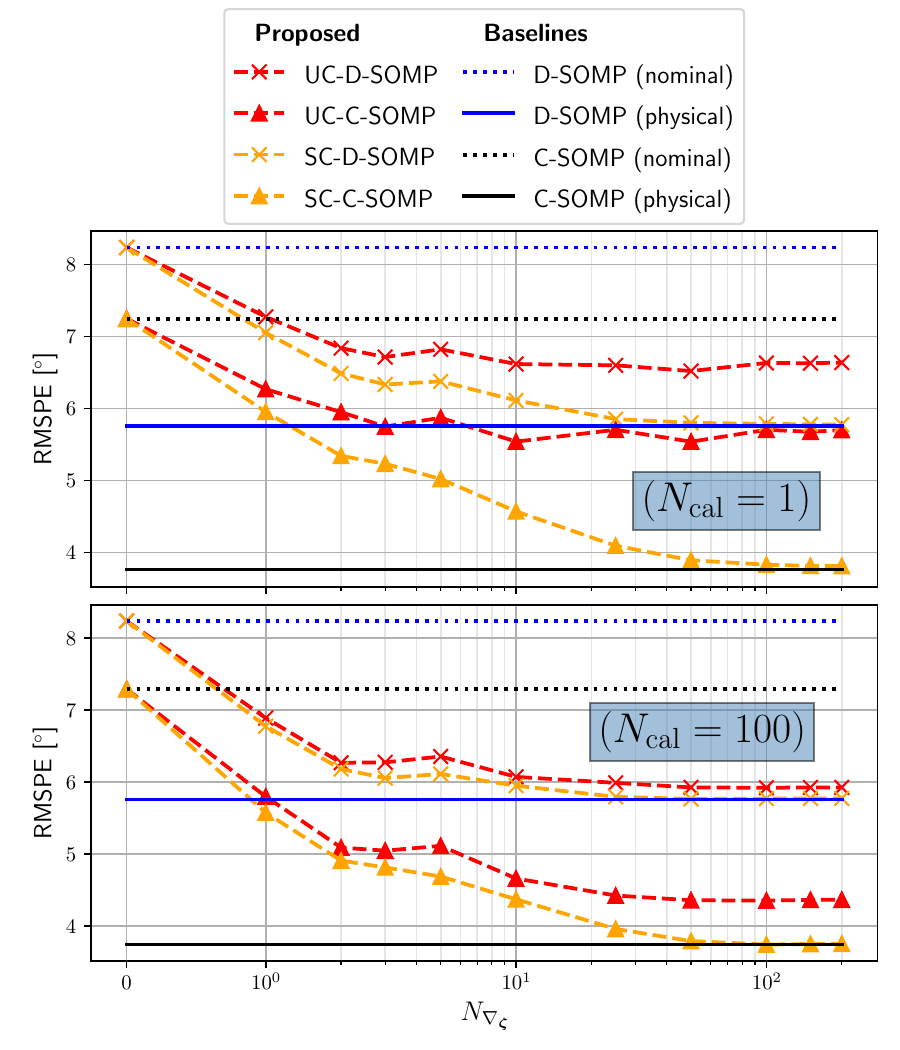}
    \caption{Performance against varying $N_{\nabla_{\bzeta}}$, for non-coherent signals. ($M=5$, $N=16$, $T=100$, $A=50$, \qty{30}{dB} SNR).}
    \label{fig:multi_gd_steps}
\end{figure}

\noindent\textbf{Impact of calibration parameters.} The previous results were obtained with $N_{\textrm{cal}} = 500$ calibration samples and $N_{\nabla_{\bzeta}} = 200$ PGD iterations, which, according to Table~\ref{tab:complexity}, entail a significant calibration cost. To assess whether this cost can be reduced, Fig.~\ref{fig:multi_gd_steps} presents the estimation performance as a function of $N_{\nabla_{\bzeta}}$, for $N_{\textrm{cal}} = 1$ and $N_{\textrm{cal}} = 100$, with $N_{\nabla_{\btheta}} = 100$. First, the calibration dataset size only affects the unsupervised approach: SC-C-SOMP reaches the performance of C-SOMP with physical knowledge for both values of $N_{\textrm{cal}}$, whereas UC-C-SOMP improves with additional samples. A single annotated measurement is therefore sufficient for SC-C-SOMP, which makes its annotation requirements almost costless. Second, even with a single measurement, at most three PGD iterations are sufficient for both UC-C-SOMP and SC-C-SOMP to reach the performance of D-SOMP with physical knowledge, which highlights the computational efficiency of the proposed method. This also underlines the importance of continuous dictionaries: with $N_{\textrm{cal}} = 1$, UC-D-SOMP never reaches the performance of D-SOMP with physical knowledge, indicating imperfect calibration, whereas its continuous counterpart UC-C-SOMP matches or exceeds it, the continuous refinement compensating for the residual array mismatch.

\begin{figure}[t]
    \centering
    \includegraphics[width=.95\columnwidth]{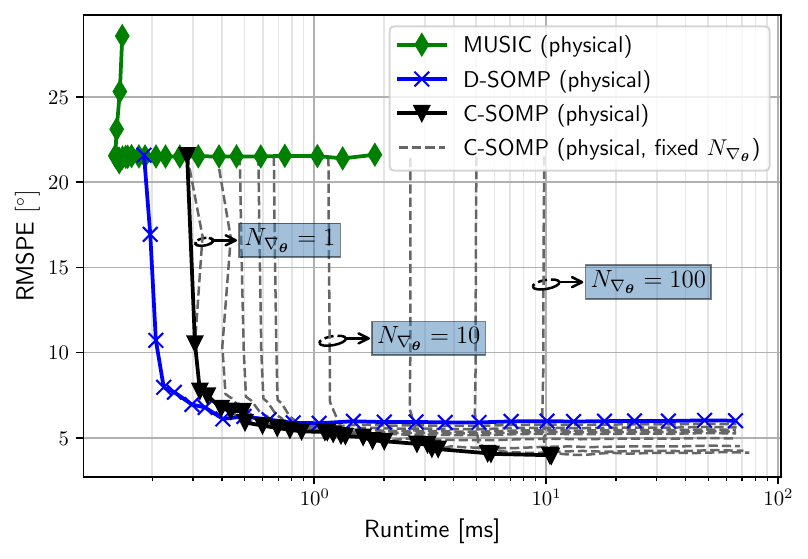}
    \caption{Performance against measured runtime, for coherent signals, $A$ being swept for each method. ($M=5$, $N=16$, $T=100$, \qty{30}{dB} SNR).}
    \label{fig:pareto_complexity}
\end{figure}

\noindent\textbf{Runtime evaluation.} Fig.~\ref{fig:pareto_complexity} reports the RMSPE against the runtime measured on one core of an NVIDIA Grace CPU, with $A$ swept for each method. For C-SOMP, each grey curve corresponds to a fixed $N_{\nabla_{\btheta}}$, and the black curve is their lower envelope. As UC-C-SOMP and SC-C-SOMP reduce to C-SOMP once calibrated, these curves also reflect their inference runtime. Consistent with Table~\ref{tab:complexity}, MUSIC is the least complex method, and the runtime of D-SOMP increases with $A$. However, MUSIC breaks down in the presence of coherent sources, while D-SOMP saturates as the grid is refined. Conversely, for each fixed $N_{\nabla_{\btheta}}$, the RMSPE of C-SOMP drops sharply at nearly constant runtime as $A$ increases, so that its runtime is mainly governed by $N_{\nabla_{\btheta}}$. Beyond approximately \qty{0.7}{ms}, C-SOMP outperforms D-SOMP at equal runtime and achieves accuracies unattainable by D-SOMP, suggesting that the condition $A_{\mathtt{C}} + \kappa N_{\nabla_{\btheta}} < A_{\mathtt{D}}$ of Remark~\ref{remark:cost} is met in practice.

\begin{figure*}[t]
    \centering
    \includegraphics[width=.62\textwidth]{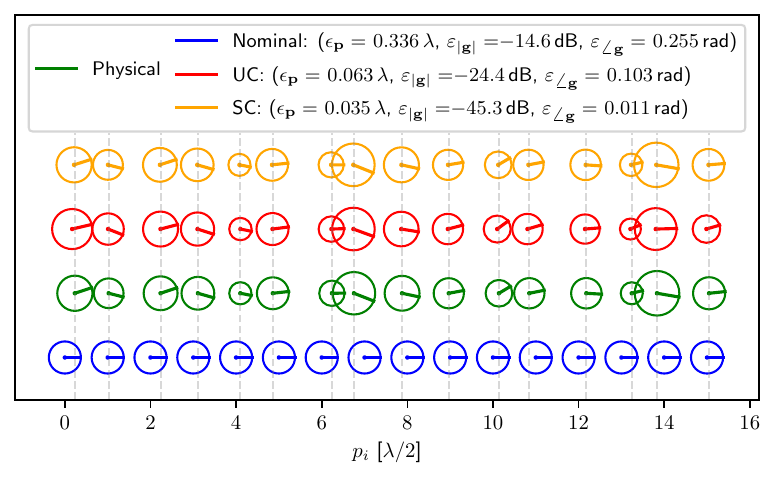}
    \caption{Learned antenna locations and gains, for coherent signals. ($M=5$, $N=16$, $T=100$, $A=50$, \qty{10}{dB} SNR, single realization, no gauge fixing).}
    \label{fig:calibration_circles}
\end{figure*}

\noindent\textbf{Analysis of learned parameters.} Previous results assessed calibration indirectly, through the RMSPE of the calibrated methods matching that of the baselines with physical array knowledge. Fig.~\ref{fig:calibration_circles} instead presents the learned antenna locations and gains for a single realization: each circle is centered on an antenna location, its radius represents the gain magnitude, and the angle of its segment with the array axis represents the gain phase. The location error is measured by $\epsilon_{\bp} = \norm{\bp - \hat{\bp}}{2}/\lambda$, the gain magnitude error by $\varepsilon_{\abs{\bg}} = 10\log_{10}(\norm{\abs{\bg}-\abs{\hat{\bg}}}{2}^2/\norm{\bg}{2}^2)$, and the gain phase error by $\varepsilon_{\angle\bg} = \norm{\angle\bg - \angle\hat{\bg}}{2}/\sqrt{N}$. Even for coherent signals at \qty{10}{dB} SNR, the parameters learned by both UC and SC closely match the physical ones, with a substantial reduction in all errors relative to the nominal array. Additionally, compared to UC for this realization, SC yields a nearly twofold reduction in the location error, improves the gain magnitude error by around \qty{20}{dB}, and reduces the gain phase error by a factor of around $10$.

\begin{figure*}[t]
    \centering
    \includegraphics[width=.96\textwidth]{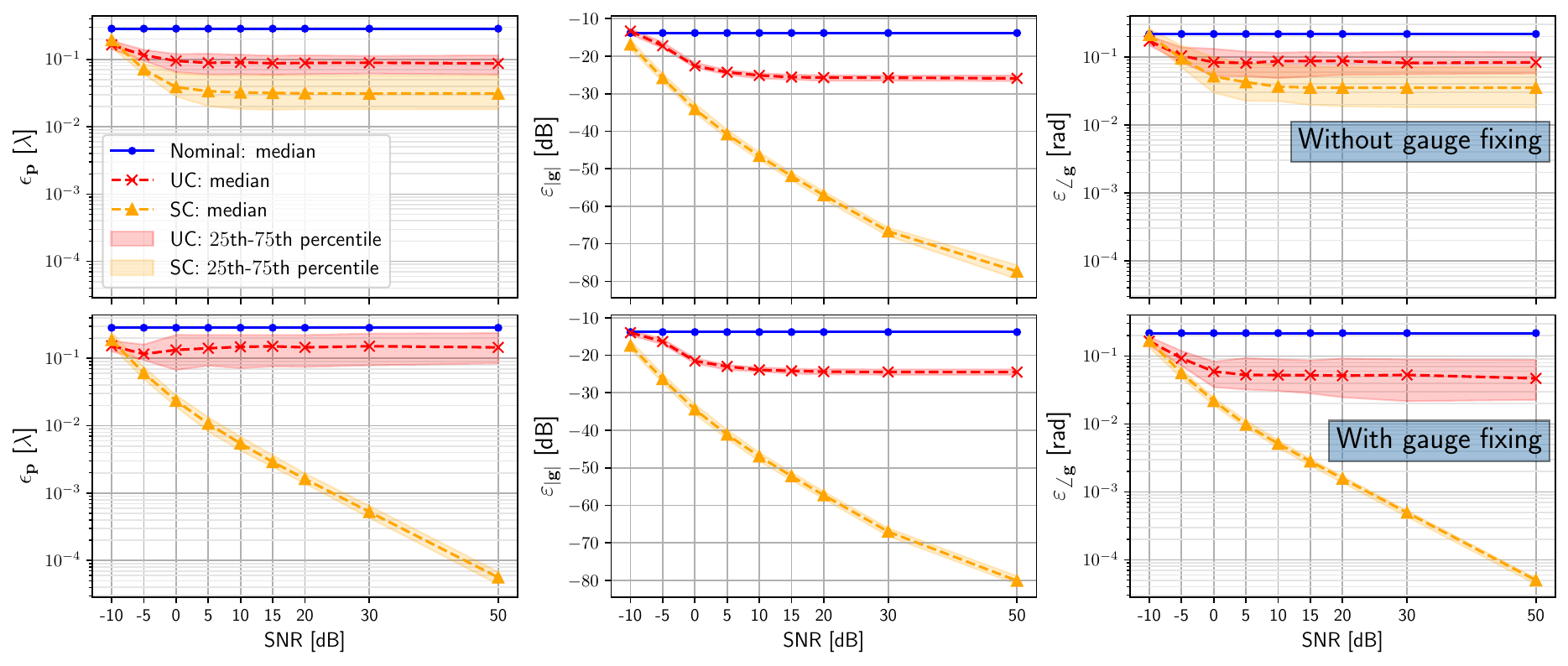}
    \caption{Calibration errors $\epsilon_{\bp}$, $\varepsilon_{\abs{\bg}}$ and $\varepsilon_{\angle \bg}$ against SNR, for coherent signals, without (top) and with (bottom) gauge fixing. ($M=5$, $N=16$, $T=100$, $A=50$).}
    \label{fig:calibration_metrics}
\end{figure*}

Fig.~\ref{fig:calibration_metrics} reports the median errors and interquartile ranges against the SNR, the errors of the nominal array serving as a reference. The top row corresponds to the proposed UC and SC methods, whereas the bottom row additionally assumes the locations and gains of the first two antennas to be known, so that only the remaining $N-2$ antenna parameters are learned. As discussed in Remark~\ref{remark:residual_ambiguity}, this prior fixes all parameters of $\mcG$, and is hence referred to as \textit{gauge fixing}. Without gauge fixing, the location and phase errors of both methods saturate at high SNR. In contrast, the gain magnitudes are unaffected by $\mcG$, so that their error directly reflects the calibration accuracy: it keeps decreasing for SC, whereas it saturates for UC, owing to the errors of the frozen DoAs estimated with the nominal array. In the SC case, the known DoAs already fix the dilation and offset, so that only the translation and global phase ambiguities remain. As expected, gauge fixing resolves them: all SC errors decrease with the SNR in the bottom row of Fig.~\ref{fig:calibration_metrics}. In contrast, the UC errors still saturate with gauge fixing, the location error even increasing. This can be explained by the dilation and offset imposed by the reference antennas conflicting with those inherited from the frozen DoAs, the remaining learned antenna parameters absorbing this mismatch. Nevertheless, the residual errors of UC without gauge fixing remain small and, as shown in previous experiments, do not prevent it from reaching the estimation performance of the baselines with physical array knowledge, which motivates not relying on reference antennas in the proposed method.

\section{Conclusion and future work}\label{sec:conclusion}

This paper investigated the joint DoA estimation and array calibration problem following the model-based machine learning paradigm. It was first shown that sparse recovery techniques possess intrinsic advantages over subspace methods, in particular their ability to perform accurate DoA estimation in the presence of coherent sources. However, sparse recovery methods are subject to imperfect knowledge of the array manifold parametrization and to a finite set of dictionary atoms, stemming from the discretization of the angular domain, both leading to estimation performance loss. This paper proposed to address both issues by jointly learning the HWIs and considering a continuous dictionary, which was formalized as a specific optimization problem. A solving procedure was then proposed, in which an unsupervised calibration stage estimates the array parametrization offline from a set of measurements, and an online estimation stage refines initial on-grid estimates over a continuous dictionary, thereby relaxing both assumptions.

Extensive experiments exposed the performance of the proposed method. It was shown that the unsupervised calibration method is able to learn the actual impaired antenna parameters with very low residual errors, even in challenging conditions with relatively high sensing noise and coherent signals. From a DoA estimation perspective, the proposed method was shown to outperform discrete sparse recovery baselines, to reach the performance of baselines with perfect impairment knowledge, and to remain effective for fully coherent sources, very low numbers of snapshots, and challenging HWI conditions. Additionally, the supervised calibration variant was shown to reach the estimation performance of a fully calibrated baseline from a single annotated measurement, in the non-coherent setting, reducing the labeling requirement to a negligible cost. 

Future work includes extending the proposed approach to mutual coupling, handling an unknown number of sources, and considering both near-field and wideband signals. Additionally, the residual dilation and offset inherited by unsupervised calibration from the nominal DoA estimates could be characterized and mitigated using weaker priors than fully known reference antennas.

\bibliographystyle{IEEEtran}
\bibliography{biblio.bib}

\appendices

\section{Proof of Proposition~\ref{prop:diffMUSIC_principal_angles}}\label{app:proof_prop_diffMUSIC}

Given two subspaces $\mcX, \mcY \subset \mbbC^N$ with orthonormal bases $\bU_{\mcX}$, $\bU_{\mcY}$, their $m = \min\left(\dim\mcX,\dim\mcY\right)$ principal angles $\Theta\left(\mcX,\mcY\right) \in \left[0,\pi/2\right]^m$ satisfy~\cite[Theorem~2.1]{ZhuKnyazev2013}:
\begin{equation}
    \cos\left(\Theta\left(\mcX,\mcY\right)\right) = \bsigma\left(\bU_{\mcY}^\herm\bU_{\mcX}\right),\label{eq:sing_values}
\end{equation}
where $\bsigma\left(\cdot\right)$ maps a matrix to the vector of its singular values $\sigma_k\left(\cdot\right)$, sorted in decreasing order, with $\sigma_{\min}\left(\cdot\right)$ and $\sigma_{\max}\left(\cdot\right)$ the smallest and largest. The squared chordal distance between $\mcX$ and $\mcY$ is $d_c^2\left(\mcX,\mcY\right) = \sum_{k=1}^{m}\sin^2\left(\Theta\left(\mcX,\mcY\right)_k\right)$.

For ease of notation, let $\bA = \bA_{\bzeta}\left(\btheta\right)$, and let $\bA = \bU_{\mcA}\bR$ be its thin QR decomposition, $\bU_{\mcA}$ being an orthonormal basis of $\mcA_{\bzeta,\btheta}$ and $\bR \in \mbbC^{M\times M}$ being invertible. As $\bA^\herm\bA = \bR^\herm\bR$, $\bR$ and $\bA$ share the same singular values. Denoting $\bB = \hat{\bU}_{\mathsf{N}}^\herm\bU_{\mcA}$, one has $\lVert\hat{\bU}_{\mathsf{N}}^\herm\bA\rVert^2_{\mathsf{F}} = \tr{\bB\bR\bR^\herm\bB^\herm}$, and $\sigma_{\min}^2\left(\bA\right)\bI_M \preceq \bR\bR^\herm \preceq \sigma_{\max}^2\left(\bA\right)\bI_M$ yields:
\begin{equation}
    \sigma_{\min}^2\left(\bA\right)\norm{\bB}{\mathsf{F}}^2 \leq \lVert\hat{\bU}_{\mathsf{N}}^\herm\bA\rVert_{\mathsf{F}}^2 \leq \sigma_{\max}^2\left(\bA\right)\norm{\bB}{\mathsf{F}}^2.
\end{equation}
Since $M \leq N-M$, $\mcA_{\bzeta,\btheta}$ and $\hat{\mcU}_{\mathsf{N}}$ have $M$ principal angles, hence:
\begin{align}
    \norm{\bB}{\mathsf{F}}^2 &= \sum_{k=1}^{M}\sigma_k^2\left(\bB\right) \overset{\textnormal{Eq.}~\eqref{eq:sing_values}}{=} \sum_{k=1}^{M}\cos^2\left(\Theta\left(\mcA_{\bzeta,\btheta},\shat{\mcU}_{\mathsf{N}}\right)_k\right)\\
    &= M - d_c^2\left(\mcA_{\bzeta,\btheta},\shat{\mcU}_{\mathsf{N}}\right),
\end{align}
which establishes Eq.~\eqref{eq:chordal_bounds}. Finally, the steering vectors being unit-norm, their orthogonality yields $\bA^\herm\bA = \bI_M$, so that $\sigma_{\min}\left(\bA\right) = \sigma_{\max}\left(\bA\right) = 1$ and Eq.~\eqref{eq:chordal_bounds} holds with equality.

\end{document}